%% file: main.tex
\documentclass[10pt,twocolumn,letterpaper]{article}

\usepackage[accsupp]{axessibility}

\usepackage{cvpr}              

\usepackage{booktabs}
\usepackage{multirow}
\usepackage{color, colortbl}
\usepackage{subcaption}
\usepackage{tabu}
\usepackage{pifont}
\usepackage{makecell}
\usepackage{paralist}
\usepackage{threeparttable}
\usepackage{enumitem}
\usepackage{hhline}
\usepackage{scalerel,xparse}

\definecolor{Gray}{gray}{0.5}
\definecolor{LGray}{gray}{0.9}
\definecolor{darkblue}{RGB}{94,110,186}
\definecolor{darkGreen}{RGB}{92, 148, 110}
\definecolor{myblue}{RGB}{14, 121, 178}
\definecolor{myred}{RGB}{192, 0, 0}

\newcommand{\blue}[1]{\textcolor{blue}{#1}}

\newcommand{\gray}[1]{\textcolor{gray}{#1}}

\newcommand{\darkGreen}[1]{\textcolor{darkGreen}{#1}}

\newcommand{\darkblue}[1]{\textcolor{darkblue}{#1}}

\newcommand{\cmark}{\ding{51}}%
\newcommand{\xmark}{\ding{55}}%

\NewDocumentCommand\ice{}{
    \scalerel*{
        \includegraphics{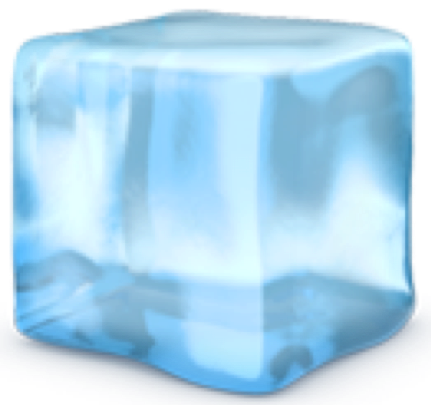}
    }{X}
}

\NewDocumentCommand\fire{}{
    \scalerel*{
        \includegraphics{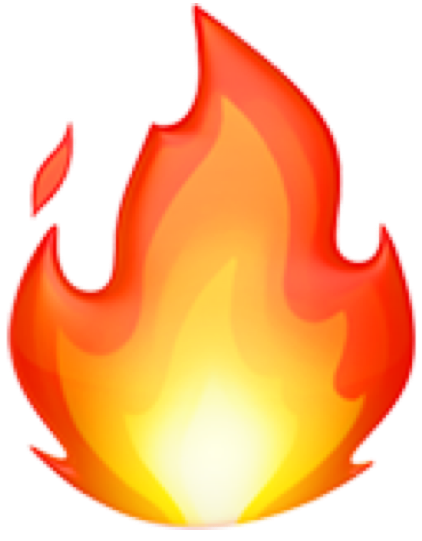}
    }{10cm}
}

\NewDocumentCommand\image{}{
    \scalerel*{
        \includegraphics{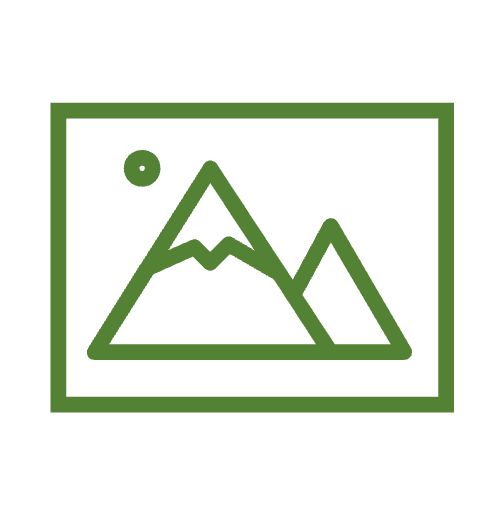}
    }{14cm}
}

\NewDocumentCommand\video{}{
    \scalerel*{
        \includegraphics{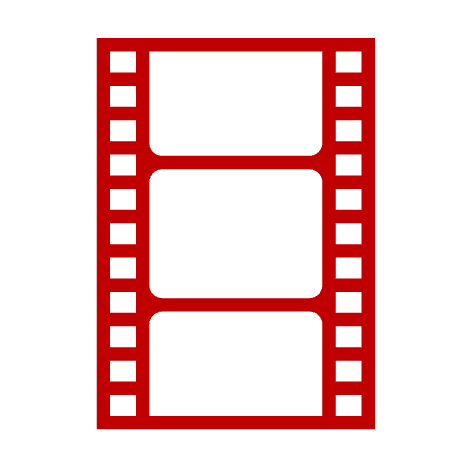}
    }{10cm}
}

\def\ModelName{VideoChat2}
\def\BenchName{MVBench}

\newcommand\blfootnote[1]{%
  \begingroup
  \renewcommand\thefootnote{}\footnote{#1}%
  \addtocounter{footnote}{-1}%
  \endgroup
}

\input{preamble}

\definecolor{cvprblue}{rgb}{0.21,0.49,0.74}
\usepackage[pagebackref,breaklinks,colorlinks,citecolor=cvprblue]{hyperref}

\title{MVBench: A Comprehensive Multi-modal Video Understanding Benchmark}

\author{
    Kunchang Li$^{1,2,3\spadesuit}$\quad
    Yali Wang$^{1,3\heartsuit}$\quad
    Yinan He$^{3}$\quad
    Yizhuo Li$^{4,3\spadesuit}$\quad
    Yi Wang$^{3}$\quad
    Yi Liu$^{1,2,3\spadesuit}$\\
    Zun Wang$^{3}$\quad
    Jilan Xu$^{5,3\spadesuit}$\quad
    Guo Chen$^{6,3\spadesuit}$\quad
    Ping Luo$^{4,3}$\quad
    Limin Wang$^{6,3\heartsuit}$\quad
    Yu Qiao$^{3,1\heartsuit}$\vspace{0.2em}\\
    \footnotesize{$^1$Shenzhen Institute of Advanced Technology, Chinese Academy of Sciences}
    \quad $^2$University of Chinese Academy of Sciences \quad $^3$Shanghai AI Laboratory\\
    \footnotesize{$^4$The University of Hong Kong\quad
    $^5$Fudan University\quad
    $^6$State Key Laboratory for Novel Software Technology, Nanjing University}
}

\begin{document}
\maketitle
\input{sec/0_abstract}   
\vspace{-0.3cm} 
\input{sec/1_intro}
\input{sec/2_related_works}
\input{sec/3_mvpbench}
\input{sec/3_mvpchat}

\input{sec/4_experiments}

\input{sec/5_conclusion}
\input{sec/6_acknowledgement}
\input{sec/X_suppl}
\clearpage
\newpage
{
    \small
    \bibliographystyle{ieeenat_fullname}
    \bibliography{main}
}


\end{document}

%% file: preamble.tex
\usepackage[dvipsnames]{xcolor}
\newcommand{\red}[1]{{\color{red}#1}}


%% file: sec/0_abstract.tex
\begin{abstract}

With the rapid development of Multi-modal Large Language Models (MLLMs), 
a number of diagnostic benchmarks have recently emerged to evaluate the comprehension capabilities of these models.
However,
most benchmarks predominantly assess spatial understanding in the static image tasks,
while 
overlooking temporal understanding in the dynamic video tasks.
To alleviate this issue,
we introduce a comprehensive \textbf{M}ulti-modal \textbf{V}ideo understanding \textbf{Bench}mark,
namely \textbf{\BenchName},
which covers \textbf{20} challenging video tasks that cannot be effectively solved with a single frame.
Specifically,
we first introduce a novel static-to-dynamic method to define these temporal-related tasks.
By transforming various static tasks into dynamic ones, 
we enable the systematic generation of video tasks that require a broad spectrum of temporal skills, ranging from perception to cognition.
Then,
guided by the task definition,
we automatically convert public video annotations into multiple-choice QA to evaluate each task.
On one hand,
such a distinct paradigm allows us to build \BenchName\ efficiently,
without much manual intervention.
On the other hand,
it guarantees evaluation fairness with ground-truth video annotations,
avoiding the biased scoring of LLMs.
Moreover,
we further develop a robust video MLLM baseline,
\textit{i.e.}, 
\textbf{\ModelName},
by progressive multi-modal training with diverse instruction-tuning data.
The extensive results on our \BenchName\ reveal that,
the existing MLLMs are far from satisfactory in temporal understanding,
while our \ModelName\ largely surpasses these leading models by over \textbf{15\%} on \BenchName.
All models and data are available at \url{https://github.com/OpenGVLab/Ask-Anything}.

\end{abstract}

%% file: sec/1_intro.tex
\section{Introduction}
\label{sec:intro}

\begin{figure}[t]
    \centering
    \includegraphics[width=1.0\linewidth
    ]{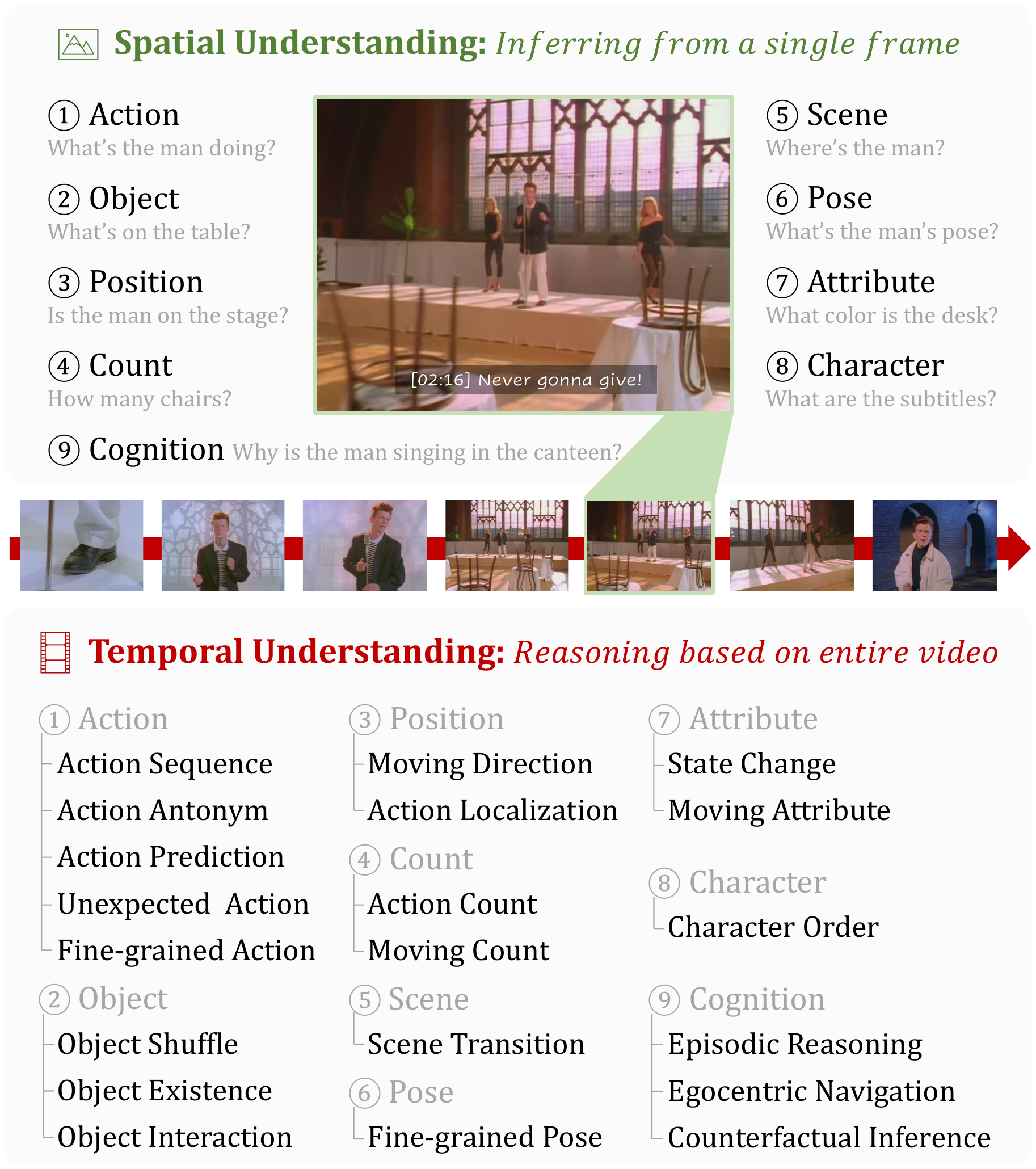}
    \vspace{-0.7cm}
    \caption{
    \textbf{Tasks of MVBench.}
    We define temporal tasks by adapting static image tasks with dynamic evolution.
    This leads to 20 challenging tasks of video understanding, 
    which cannot be effectively solved within a single frame.
    For example, 
    ``\textit{position}'' in an image can be converted into ``\textit{moving direction}'' through a video.
    }
    \label{fig:intro}
    \vspace{-0.5cm}
\end{figure}

In the past few years,
\blfootnote{$\spadesuit$ Interns at Shanghai AI Laboratory. $\heartsuit$ Corresponding authors.}
Multi-modal Large Language Models (MLLMs)~\cite{gpt4v,flamingo,blip2,palme,minigpt4,llava,kosmos,videochat} 
have gradually driven the advance in vision-language learning, 
by plugging visual encoders within various pretrained LLMs~\cite{devlin2018bert,palm,chatgpt,llama1,vicuna}. 
With such a fast development,
there is a natural question: 
\textit{How can we evaluate the comprehension capabilities of these MLLMs?}
Such assessment is vital to confirm their design effectiveness and further improve them for a broader understanding of open-world multi-modalities.

In response to this need,
a number of benchmarks have been launched~\cite{pope,lvlm_ehub,mme,mmbench,mmvet},
by evaluating MLLMs with Question Answering (QA) formulation of various perception tasks.
However,
most of these benchmarks primarily concentrate on image-based understanding,
where
all the questions are designed for spatial perception in the static images,
\textit{e.g.},
``\textit{Is the man on the stage?}'',
as shown in Fig. \ref{fig:intro}.
Hence,
they suffer from difficulty in assessing temporal evolution in dynamic videos,
which is critical to understanding the procedural activities in our realistic world.
Recently,
several attempts have tried to evaluate MLLMs on temporal perception in videos~\cite{seedbench,funqa,perception_test,videochatgpt}. 
But they either work on the very basic video tasks (\textit{e.g.}, action recognition and prediction in SEED-Bench~\cite{seedbench}), 
or focus on the particular domains (\textit{e.g.}, surprising comprehension in FunQA~\cite{funqa}) and restricted scenes (\textit{e.g.}, indoor scenes in Perception Test~\cite{perception_test}).
As a result,
it is limited to leverage these benchmarks to make a comprehensive evaluation on the temporal understanding skills of MLLMs.
Besides,
they are collected with labor-intensive annotations,
leading to expensive manual intervention.
To tackle these problems, 
we propose a \textbf{M}ulti-modal \textbf{V}ideo understanding \textbf{Bench}mark (\textbf{\BenchName}),
which aims at comprehensively evaluating the temporal perception capabilities of MLLMs in the open world.
Compared to these existing benchmarks above,
there are two distinct designs in our \BenchName.

First,
we introduce a novel static-to-dynamic method to systematically define temporal-related tasks,
by adapting static image tasks with dynamic evolution.
This leads to \textbf{20} challenging tasks of video understanding in the \BenchName,
which covers a wide range of temporal understanding skills from perception to cognition. 
Specifically,
we use static image tasks in the previous multi-modal benchmarks~\cite{mme,mmbench} as definition reference.
Then,
we augment the question of these static tasks with temporal context in the video,
\textit{e.g.},
the \textit{position} task in the image can be flexibly converted into the \textit{moving-direction} task in the video 
(``\textit{Is the man on the stage?}'' $\rightarrow$ ``\textit{What direction is the man moving?}'') in Fig. \ref{fig:intro}. 
In this case,
we can effectively convert all these static tasks into the corresponding dynamic tasks,
which cannot be solved without reasoning on the whole video.

Second,
guided by the task definition,
we design an automatic annotation paradigm to generate multiple-choice QAs for each task,
by converting \textbf{11} public video benchmarks with LLMs.
On one hand,
it can largely reduce the cost of expensive human annotations.
On the other hand,
these 11 benchmarks cover various complex domains and diverse scenes,
ranging from first-person to third-person perspectives, 
and from indoor to outdoor environments.
Hence,
our \BenchName\ is a preferable choice to evaluate the general capability of MLLMs for open-world temporal understanding.
More importantly,
these benchmarks provide the ground truth for \BenchName\,
which guarantees evaluation fairness and accuracy,
avoiding biased scoring of LLMs~\cite{funqa,videochatgpt}.

Finally,
we make a thorough evaluation of various well-known MLLMs on our \BenchName.  
Surprisingly,
these state-of-the-art image and video MLLMs are far from satisfactory,
in terms of temporal perception and cognition.
This further motivates us to develop a strong video MLLM baseline,
namely \textbf{\ModelName},
by bridging LLM with a powerful vision foundation model~\cite{umt}.
Subsequently,
we introduce a progressive training paradigm with a wide spectrum of multi-modal instructions,
allowing effective alignment between video and language.
The evaluations show that,
our \ModelName\ significantly surpasses the top-performing VideoChat~\cite{videochat} by over \textbf{15\%} accuracy on \BenchName,
and also achieves the new state-of-the-art results on video conversation~\cite{videochatgpt} and zero-shot QA benchmarks~\cite{msrvtt_qa,activitynet_qa}. 
All the models and data are publicly available,
in order to pave the path to general video understanding.

%% file: sec/2_related_works.tex
\section{Related Works}

\noindent\textbf{MLLM.}
Building upon the significant achievements of Large Language Models (LLMs)~\cite{devlin2018bert,t5,gpt3,Wei2021FinetunedLM,palm}, scholarly interest has increasingly shifted towards the exploration and development of Multi-modal Large Language Models (MLLMs). 
This shift aims to augment multi-modal understanding and generation capabilities. 
Groundbreaking MLLMs such as Flamingo~\cite{flamingo} and PaLM-E~\cite{palme} have seamlessly fused text and vision, setting precedence with their outstanding performances across a range of multi-modal tasks~\cite{vqa,fickr,msrvtt,okvqa}.
The recent open-sourcing of LLMs~\cite{llama1,llama2,vicuna,glm,internlm} further accelerates the emergence of public MLLMs~\cite{minigpt4,llava,mmgpt}. 
Notable examples such as LLaVA~\cite{llava}, MiniGPT-4~\cite{minigpt4}, and InstructBLIP~\cite{instructblip} have contributed by proposing a series of visual instruction-tuning data.
Venturing beyond text and static images, several studies have begun harnessing video modality~\cite{videochat,videochatgpt,videollama,valley}, tapping into the vast potential of LLMs for video comprehension tasks~\cite{msrvtt_qa,activitynet_qa,msvd}. 
Innovations like VideoChat~\cite{videochat}, VideoChatGPT~\cite{videochatgpt}, and Valley~\cite{valley} utilize ChatGPT to generate video instruction-tuning data, aiming to enhance instruction-following capabilities.
In the \ModelName, we aim to critically examine the fundamental temporal understanding capabilities of MLLMs, providing valuable design insights for more robust video MLLMs.

\input{tables/task_dimensions}

\noindent\textbf{Benchmark.}
Traditional Vision-Language (VL) benchmarks~\cite{k400,sth,msrvtt,nextqa,msrvtt_qa} have primarily honed in on specific capabilities like multi-modal retrieval and vision QA. 
The rise of MLLMs has catalyzed benchmarks designed for assessing integrated VL tasks. 
For example, LVLM-eHub~\cite{lvlm_ehub} provides an interactive model comparison platform through image-related queries. 
Other benchmarks such as OwlEval~\cite{mplug-owl}, MME~\cite{mme}, SEED-Bench~\cite{seedbench}, MM-Vet~\cite{mmvet}, and MMBench~\cite{mmbench} underscore comprehensive VL skills, introducing evaluation metrics that transcend mere model hierarchies.
Meanwhile, the video realm showcased benchmarks like Perception Test~\cite{perception_test}, examining multi-modal video perception and reasoning, and VideoChatGPT~\cite{videochatgpt} quantifies the capability of dialogue generation from video inputs.
FunQA~\cite{funqa} pushes video reasoning limits via counter-intuitive and humorous content.
In contrast to the existing benchmarks,
MVBench sets itself apart by covering a wide range of temporal tasks, 
emphasizing temporally-sensitive videos and efficient use of public annotations, 
and conducting comprehensive evaluations of MLLMs' temporal understanding.

%% file: tables/task_dimensions.tex
\begin{table*}[tp]
    \centering
    \setlength\tabcolsep{4pt}
    \resizebox{1.0\textwidth}{!}{
        \begin{tabular}{c|c|c|l}
        \Xhline{1.0pt}
        \textbf{Spatial} & \textbf{Temporal} & \textbf{Source} & \textbf{Example} \\
        \Xhline{1.0pt}
        \multirow{11}{*}{\textbf{Action}} & Action & \multirow{2}{*}{STAR} & \cellcolor{gray!5}{\textit{\darkblue{What happened after the person took the food?}}} \\
        ~ & Sequence & ~ & \cellcolor{gray!5}{(A) Ate the medicine. (B) Tidied up the blanket. (C) Put down the cup/glass/bottle. (D) Took the box.} \\
        \hhline{~|-|-|-}
        ~ & Action & \multirow{2}{*}{STAR} & \cellcolor{gray!5}{\textit{\darkblue{What will the person do next?}}} \\
        ~ & Prediction & ~ & \cellcolor{gray!5}{(A) Put down the pillow. (B) Open the door. (C) Take the book. (D) Open the closet/cabinet.} \\
        \hhline{~|-|-|-}
        ~ & Action & \multirow{2}{*}{PAXION\red{$\ddag$}} & \cellcolor{gray!5}{\textit{\darkblue{Which one of these descriptions correctly matches the actions in the video?}}}  \\
        ~ & Antonym & ~ & \cellcolor{gray!5}{(A) not sure (B) scattering something down (C) piling something up} \\
        \hhline{~|-|-|-}
        ~ & Fine-grained & \multirow{2}{*}{MiT V1\red{$\ddag$}} & \cellcolor{gray!5}{\textit{\darkblue{What is the action performed by the person in the video?}}}  \\
        ~ & Action & ~ & \cellcolor{gray!5}{(A) watering (B) leaking (C) pouring (D) planting} \\
        \hhline{~|-|-|-}
        ~ & \multirow{3}{*}{\makecell[c]{Unexpected\\Action}} & \multirow{3}{*}{FunQA\red{$\ddag$}} & \cellcolor{gray!5}{\textit{\darkblue{What unexpected event contributes to the humor in the video?}}}  \\
        ~ & ~ & ~ & \cellcolor{gray!5}{(A) The man left without dancing. (B) Two women hugged each other at the end.} \\
        ~ & ~ & ~ & \cellcolor{gray!5}{(C) The man finally danced with the woman. (D) Two men hugged each other unexpectedly.} \\
        \hline
        \multirow{5}{*}{\textbf{Object}} & Object Existence & CLEVRER & \cellcolor{gray!5}{\textit{\darkblue{Are there any moving green objects when the video ends?}} (A) not sure (B) yes (C) no} \\
        \hhline{~|-|-|-}
        ~ & Object Interaction & STAR & \cellcolor{gray!5}{\textit{\darkblue{Which object was tidied up by the person?}} (A) broom (B) cabinet (C) blanket (D) table} \\
        \hhline{~|-|-|-}
        ~ & \multirow{3}{*}{\makecell[c]{Object\\Shuffle}} & \multirow{3}{*}{\makecell[c]{Perception\\Test}} & \cellcolor{gray!5}{\textit{\darkblue{Where is the hidden object at the end of the game from the person's point of view?}}} \\
        ~ & ~ & ~ & \cellcolor{gray!5}{(A) Under the first object from the left. (B) Under the third object from the left.}  \\
        ~ & ~  & ~ & \cellcolor{gray!5}{(C) Under the second object from the left.} \\
        \hline
        \multirow{5}{*}{\textbf{Position}} & Moving & \multirow{2}{*}{CLEVRER\red{$\ddag$}} & \cellcolor{gray!5}{\textit{\darkblue{What direction is the cyan sphere moving within the video?}}} \\
        ~ &  Direction & ~ & \cellcolor{gray!5}{(A) The object is stationary. (B) Up and to the right. (C) Down and to the left. (D) Down and to the right.} \\
        \hhline{~|-|-|-}
        ~ & \multirow{3}{*}{\makecell[c]{Action\\Localization}} & \multirow{3}{*}{\makecell[c]{Charades-\\STA}\red{$\ddag$}} & \cellcolor{gray!5}{\textit{\darkblue{During which part of the video does the action `person sitting on a couch' occur?}}} \\
        ~ & ~ & ~ & \cellcolor{gray!5}{(A) In the middle of the video. (B) At the end of the video.} \\
        ~ & ~ & ~ & \cellcolor{gray!5}{(C) Throughout the entire video. (D) At the beginning of the video.} \\
        \hline
        \multirow{3}{*}{\textbf{Scene}} & \multirow{3}{*}{\makecell[c]{Scene\\Transition}} & \multirow{3}{*}{MoVQA\red{$\ddag$}} & \cellcolor{gray!5}{\textit{\darkblue{What's the right option for how the scenes in the video change?}}} \\
        ~ & ~ & ~ & \cellcolor{gray!5}{(A) From the reception desk to the conference room. (B) From the kitchen to the dining area.} \\
        ~ & ~ & ~ & \cellcolor{gray!5}{(C) From the server room to the control center. (D) From the classroom to the library.} \\
        \hline
        \multirow{2}{*}{\textbf{Count}} & Action Count & \small{Perception Test} & \cellcolor{gray!5}{\textit{\darkblue{How many times did the person launch objects on the table?}} (A) 3 (B) 2 (C) 4} \\
        \hhline{~|-|-|-}
        ~ & Moving Count & CLEVRER & \cellcolor{gray!5}{\textit{\darkblue{How many metal objects exit the scene?}} (A) 2 (B) 3 (C) 1 (D) 0} \\
        \hline
        \multirow{2}{*}{\textbf{Attribute}} & Moving Attribute & CLEVRER & \cellcolor{gray!5}{\textit{\darkblue{What shape is the moving object when the video begins?}} (A) cylinder (B) sphere (C) cube} \\
        \hhline{~|-|-|-}
        ~ & State Change & \small{Perception Test} & \cellcolor{gray!5}{\textit{\darkblue{Is the lighting device on at any point?}} (A) yes (B) I don't know (C) no} \\
        \hline
        \textbf{Pose} & Fine-grained Pose & \small{NTU RGB+D\red{$\ddag$}} & \cellcolor{gray!5}{\textit{\darkblue{What is the pose performed by the person in the video?}} (A) pick up (B) sit down (C) drop (D) stand up} \\
        \hline
        \textbf{Character} & Character Order & \small{Perception Test} & \cellcolor{gray!5}{\textit{\darkblue{What letter did the person write first on the paper?}} (A) l (B) v (C) e} \\
        \hline
        \multirow{9}{*}{\textbf{Cognition}} & \multirow{2}{*}{\makecell[c]{Egocentric\\Navigation}} & \multirow{2}{*}{VLN-CE\red{$\ddag$}} & \cellcolor{gray!5}{\textit{\darkblue{For an agent following instruction: ``Go left through the door.'' What is the next action it should take?}}}  \\ 
        ~ & ~ & ~ & \cellcolor{gray!5}{(A) Turn left and move forward (B) Move forward (C) Stop (D) Turn right and move forward.}  \\ 
        \hhline{~|-|-|-}
        ~ & \multirow{3}{*}{\makecell[c]{Episodic\\Reasoning}} & \multirow{3}{*}{TVQA} & \cellcolor{gray!5}{\textit{\darkblue{Why did Castle dress like a fairy when he was speaking to Emily?}}} \\
        ~ & ~ & ~ & \cellcolor{gray!5}{(A) To get her to trust him. (B) He secretly loved fairies. (C) He lost a bet with Emily.} \\
        ~ & ~ & ~ & \cellcolor{gray!5}{(D) It was dressed like a fairy day at school. (E) Mrs Ruiz made him dress up.} \\
        \hhline{~|-|-|-}
        ~ & \multirow{3}{*}{\makecell[c]{Counterfactual\\Inference}} & \multirow{3}{*}{CLEVRER} & \cellcolor{gray!5}{\textit{\darkblue{Which of the following will happen if the cylinder is removed?}}} \\
        ~ & ~ & ~ & \cellcolor{gray!5}{(A) The cyan rubber object and the blue cube collide. (B) The brown cube collides with the metal cube.}  \\
        ~ & ~ & ~ & \cellcolor{gray!5}{(C) The cyan rubber object and the metal cube collide. (D) The cyan rubber cube collides with the sphere.} \\
        \Xhline{1.0pt}
        \end{tabular}
    }
    \vspace{-0.3cm}
    \caption{
    \textbf{Task examples of \BenchName.}
    The videos are collected from the public datasets, including STAR~\cite{star}, PAXION~\cite{paxion}, Moments in Time V1~\cite{mit}, FunQA~\cite{funqa}, CLEVRER~\cite{clevrer}, Perception Test~\cite{perception_test}, Charades-STA~\cite{charades_sta}, MoVQA~\cite{movqa}, NTU RGB+D\cite{ntu_rgbd}, VLN-CE~\cite{vln_ce} and TVQA ~\cite{tvqa}.
    Tasks requiring QA generation are marked with ``\red{$\ddag$}''.
    More details can be found in Section \ref{sec:dimension}.
    }
    \label{tab:task_dimension}
    \vspace{-0.3cm}
\end{table*}

%% file: sec/3_mvpbench.tex
\section{\BenchName}
\label{sec:mvpbench}

\begin{figure*}[thp]
    \centering
    \includegraphics[width=0.95\textwidth]{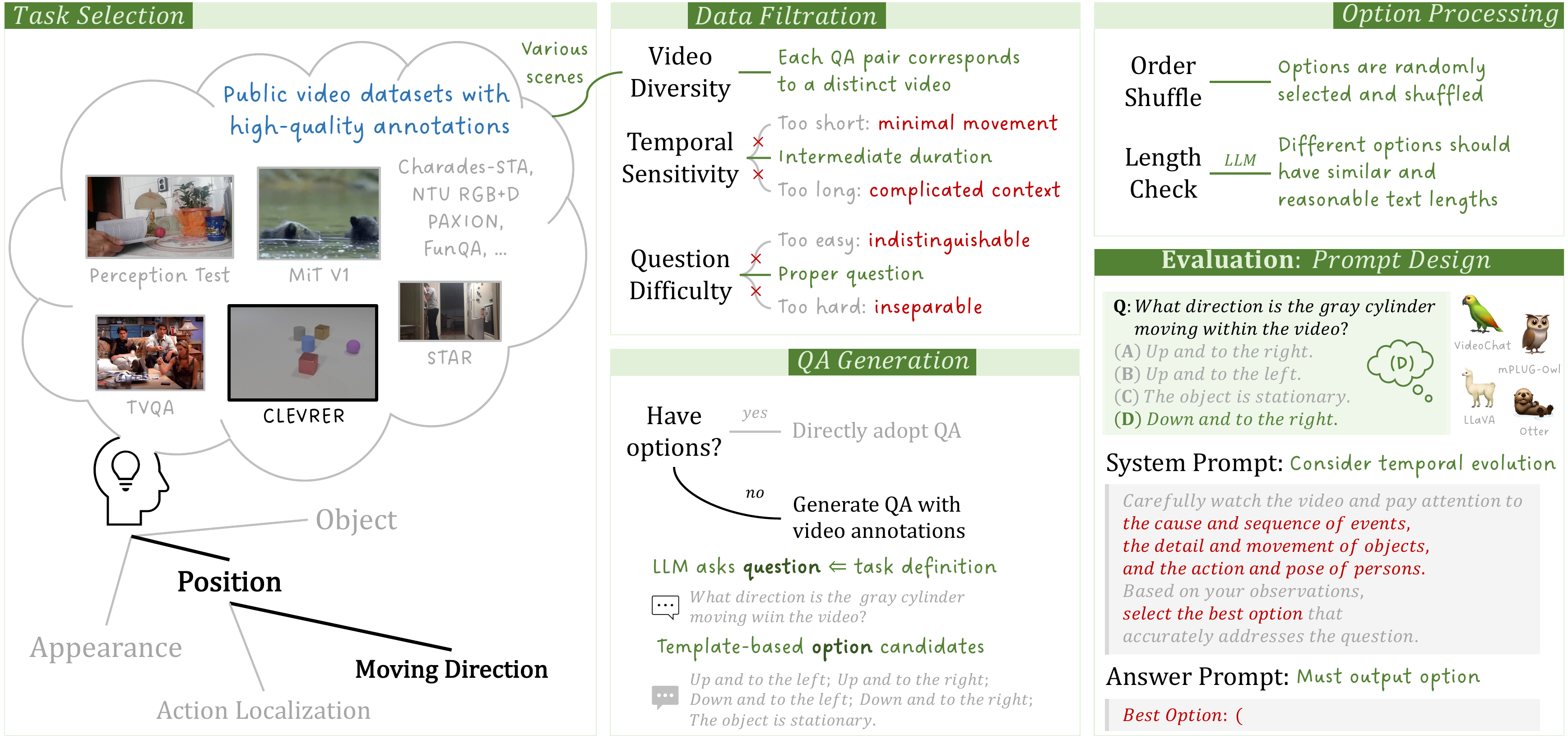}
    \vspace{-0.3cm}
    \caption{
    \textbf{Generation pipeline of \BenchName.} 
    Within public annotations, 
    data is carefully filtered and relevant multiple-choice QAs are auto-generated.
    The effective system prompt and efficient answer prompt are employed to guide MLLMs toward precise outputs.
    }
    \label{fig:pipeline}
    \vspace{-0.3cm}
\end{figure*}

In this section, 
we present our \BenchName\ in detail. 
We first design the temporal tasks in Tab. \ref{tab:task_dimension},
and then automatically generate multiple-choice QAs for evaluation in Fig. \ref{fig:pipeline}.

\subsection{Temporal Task Definition}
\label{sec:dimension}

To design the temporal tasks of \BenchName, 
we introduce a concise static-to-dynamic method by adapting static tasks with dynamic goals.
As discussed in the introduction,
most existing MLLM benchmarks~\cite{mme,mmbench} focus on spatial understanding with systematical definitions of static image tasks.
Motivated by this, 
we propose using these task definitions as references to systematically design temporal tasks, ranging from perception to cognition.
As shown in Fig. \ref{fig:intro}, 
we begin by summarizing 9 main tasks of spatial understanding from previous benchmarks. 
Then we enrich these image tasks with video context, 
creating temporal tasks that can not be effectively solved with a single image and require comprehensive video understanding. 
Finally, 
we define 20 temporal tasks as follows. 
Examples are listed in Tab. \ref{tab:task_dimension}.

\textbf{Action.}
(1) \textit{Action Sequence:} Retrieve the events occurring before or after a specific action.
(2) \textit{Action Prediction:} Infer the subsequent events based on the current actions.
(3) \textit{Action Antonym:} Distinguish the correct action from two inversely ordered actions.
(4) \textit{Fine-grained Action:} Identify the accurate action from a range of similar options.
(5) \textit{Unexpected Action:} Detect surprising actions in videos characterized by humor, creativity, or magic.
\textbf{Object.} 
(6) \textit{Object Existence:} Determine the existence of a specific object during a particular event.
(7) \textit{Object Interaction:} Identify the object that participates in a particular event.
(8) \textit{Object Shuffle:} Locate the final position of an object in an occlusion game.
\textbf{Position.}
(9) \textit{Moving Direction:} Ascertain the trajectory of a specific object's movement.
(10) \textit{Action Localization:} Determine the time period when a certain action occurs.
\textbf{Scene.} 
(11) \textit{Scene transition:} Determine how the scene transitions in the video.
\textbf{Count.}
(12) \textit{Action Count:} Calculate how many times a specific action has been performed.
(13) \textit{Moving Count:} Calculate how many objects have performed a certain action.
\textbf{Attribute.}
(14) \textit{Moving Attribute:} Determine the appearance of a specific moving object at a given moment.
(15) \textit{State Change:} Determine whether the state of a certain object changes throughout the video.
\textbf{Pose.} 
(16) \textit{Fine-grained Pose:} Identify the accurate pose category from a range of similar options.
\textbf{Character.}
(17) \textit{Character Order:} Determine the order in which the letters appear.
\textbf{Cognition.}
(18) \textit{Egocentric Navigation:} Forecast the subsequent action, based on an agent's current navigation instructions.
(19) \textit{Episodic Reasoning:} Perform reasoning on the characters, events, and objects within an episode of a TV series.
(20) \textit{Counterfactual Inference:} Consider what might happen if a certain event occurs.

\subsection{Automatic QA Generation}
\label{sec:generation}

With the guidance of temporal task definitions,
we next collect and annotate videos for each task.
Specifically,
we design an automatic QA generation paradigm in Fig. \ref{fig:pipeline},
which efficiently converts open-sourced video annotations into multiple-choice QAs for evaluating MLLMs.

\begin{figure*}[thp]
    \centering
    \includegraphics[width=0.95\textwidth]{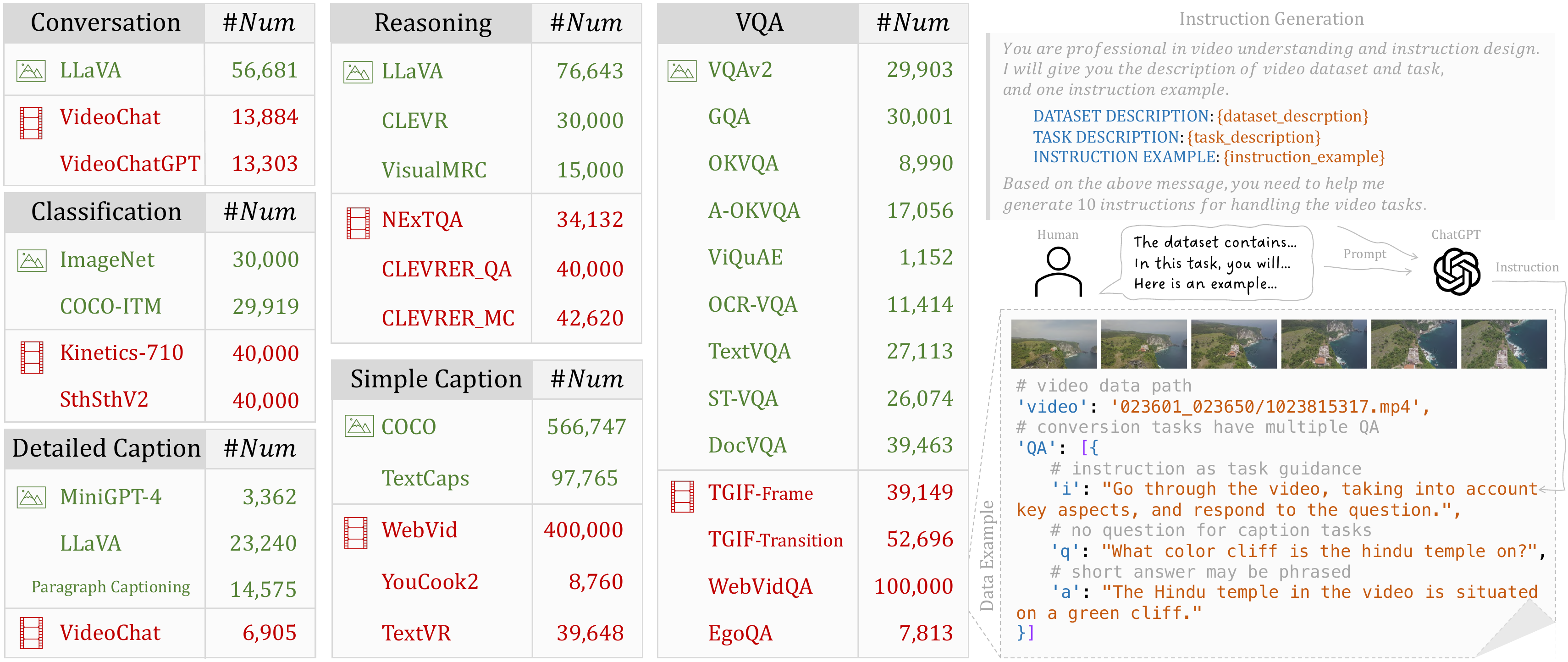}
    \vspace{-0.35cm}
    \caption{
    \textbf{Instruction-tuning data for \ModelName.}
    Co-training of \ModelName\  employs both image and video data, with instructions generated by ChatGPT~\cite{chatgpt}. 
    The resultant dataset comprises 2M samples drawn from 34 diverse datasets across 6 categories. 
    }
    \label{fig:instruction}
    \vspace{-0.3cm}
\end{figure*}

\textbf{Data Filtration.} 
To reduce the labor-intensive collection,
we propose to select videos from existing benchmarks.
\textbf{(1) Video Diversity:}
To boost video diversity,
we carefully select 11 video datasets (see Tab. \ref{tab:task_dimension}) that cover a broad spectrum of domains and scenes,
ranging from first-person to third-person perspectives, 
and from indoor to outdoor environments.
\textbf{(2) Temporal Sensitivity:}
To guarantee that each task is temporal sensitive,
we eliminate short clips which generally contain negligible motions, 
and also delete extremely long videos which often present overly complicated contexts that are hard for evaluation. 
Hence,
we select videos with intermediate duration, primarily ranging from 5s to 35s.
\textbf{(3) Question Difficulty:}
Overly simple or complex questions may lead to indistinguishable evaluations, due to similar responses.
To balance the question difficulty,
we design the selection criteria for STAR~\cite{star} and CLEVRER~\cite{clevr}.
For STAR, 
we enhance the challenge by randomly shifting the start or end points of the video clips, 
increasing the complexity of localizing specific events. 
For CLEVRER, 
we exclude questions that necessitate more than 10 conditions (\textit{e.g.,} material, and shape) for describing specific events,
thus decreasing QA difficulty.

\textbf{QA Generation.}
Considering that not all the annotations of selected datasets follow the multiple-choice QA format,
we automatically convert the video annotations into this format via LLMs.
Specifically,
we first use ChatGPT~\cite{chatgpt} to generate a question for each video,
based on the task definition.
Then,
we create the corresponding answer options as follows.
\textbf{(1) Template-Based Construction:} 
For most questions, 
we construct the option candidates directly from the ground truth annotations. 
For example,
the candidates for the \textit{Action Antonym} task contain
the \textit{correct} action, 
its \textit{opposite} action, 
and a \textit{not-sure} choice. 
In the case of the \textit{Moving Direction} task,
the option candidates consist of four directions (\textit{i.e.}, \textit{up}, \textit{down}, \textit{left}, \textit{right}) and the \textit{stationary} state.
\textbf{(2) LLM-Based Generation:}
For the \textit{Unexpected Action} task in particular, 
we leverage ChatGPT for converting open-ended QAs into multiple-choice QA with answer options.
Note that,
we use the multiple-choice format instead of the open-ended one,
for evaluation correction and fairness.
This is mainly because the open-ended answer has to be scored by LLMs or user studies,
which may either introduce evaluation bias or manual intervention.
Ultimately, 
we produce 200 multiple-choice QA pairs for each temporal understanding task.
More details of QA generation for all the tasks can be found in the appendix.

\textbf{Answer Option Processing.}
For all questions,
we randomly sample 3 to 5 answer options from the available candidates,
and shuffle the option order,
to strengthen the evaluation's robustness.
Additionally,
to prevent the common issue of answer leakage where longer options tend to be correct,
we further use LLM to guarantee that all the answer options of a question are of similar and reasonable lengths.

\subsection{Prompt Design for Evaluation}
\label{sec:evaluation}

To emphasize the temporal sensitivity of MLLMs,
we craft a detailed \textbf{system prompt} for evaluation (see the bottom right of Fig. \ref{fig:pipeline}). 
This prompt encourages MLLMs to carefully scrutinize video content to answer questions,
by paying attention to factors such as 
the actions and poses of persons, 
and the details and movements of object movements.

Moreover,
another significant challenge lies in extracting options from MLLMs' responses. 
MMBench~\cite{mmbench} attempts to match predictions with multiple option formats. 
If failed, 
it resorts to ChatGPT~\cite{chatgpt} to extract options through an intricate design. 
However, 
this way is relatively inefficient, 
yielding an alignment rate of only 87\% with humans.
In contrast,
our \BenchName\ employs a simple approach that guarantees 100\% rate in option extraction.
We enclose the options within parentheses in the questions,
and use the \textbf{answer prompt} ``\textit{Best Option}: ('' to guide MLLMs for option generation. 
Results in Tab. \ref{tab:ablation_answer_prompt} demonstrate our prompt's effectiveness on various MLLMs, 
allowing us to use accuracy as a reliable metric for evaluation.

%% file: sec/3_mvpchat.tex
\section{\ModelName }
\label{sec:mvpchat}

After building our \BenchName,
we evaluate a number of popular image and video MLLMs in Tab. \ref{tab:mvpbench}.
Surprisingly,
the existing MLLMs are far from satisfactory in temporal understanding.
To fill the gap,
we develop a robust video MLLM baseline,
which is dubbed as \textbf{\ModelName}.

\begin{figure*}[thp]
    \centering
    \includegraphics[width=0.95\textwidth]{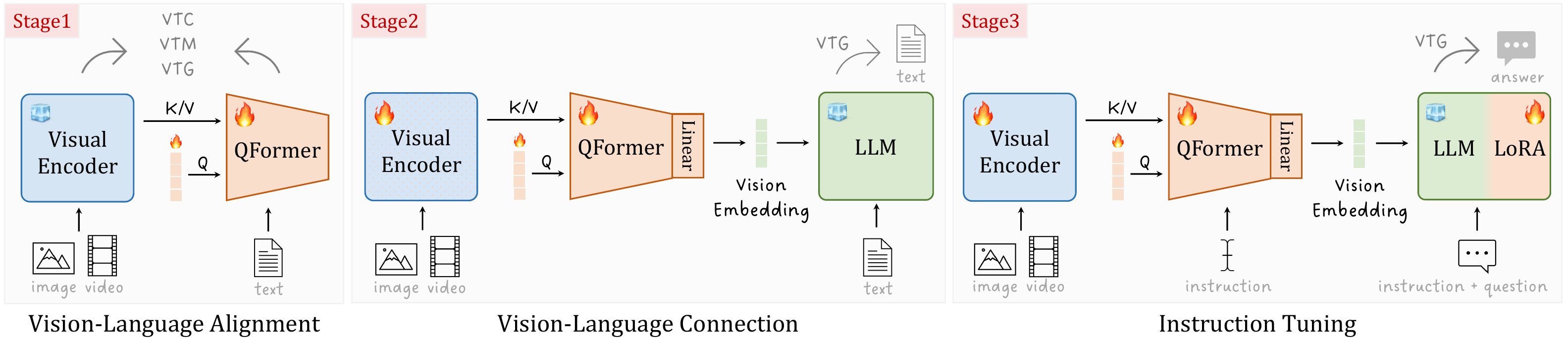}
    \vspace{-0.3cm}
    \caption{
    \textbf{Progressive multi-modal training of \ModelName.}
    Stage1 aligns UMT-L~\cite{umt}, the visual encoder, with QFormer~\cite{blip2} to efficiently compress extensive visual inputs. 
    Stage2 extends this connection to incorporate LLM, 
    while Stage3 focuses on effective instruction tuning to enhance model performance.
    The terms `\textit{instruction}', `\textit{question}' and `\textit{answer}' means `\texttt{i}', `\texttt{q}' and `\texttt{a}' of `\texttt{QA}' in Fig. \ref{fig:instruction}.
    }
    \label{fig:model}
    \vspace{-0.3cm}
\end{figure*}

\subsection{Instruction-Tuning Data}
\label{sec:instruction_data}

Primarily, 
the suboptimal performance of MLLMs can be attributed to the limited diversity in instruction-tuning data. 
To address this issue, 
we introduce the enriched data as shown in Fig. \ref{fig:instruction}, 
which comprises 2M samples from 34 distinct sources.
Following \cite{videochat,videollama},
we include both image and video data in the instruction set to improve training.

Motivated by M$^3$IT~\cite{m3it},
we reorganize all data samples in a uniform format, as shown on the bottom right of Fig. \ref{fig:instruction}. 
There are two keys involved: \{`\texttt{image}' or `\texttt{video}'\}, and \{`\texttt{QA}'\}.
The first key indicates the path to the vision data.
The second key represents a list that contains task instruction (`\texttt{i}') and question-answer(`\texttt{q}'-`\texttt{a}').
Moreover,
different from M$^3$IT, 
which requires researchers to write 10 instructions per dataset, 
we use ChatGPT to create them, 
according to \{dataset description\}, \{task description\}, and \{instruction example\} at the top right of Fig. \ref{fig:instruction}.
Consequently,
our whole instruction-tuning data set can be roughly divided into 6 categories as follows:

\textbf{(1) Conversation} 
aims at enhancing multi-turn conversational capabilities. 
We collect conversation data from LLaVA~\cite{llava} and VideoChat~\cite{videochat}.
To expand our data, 
we integrate the caption data from VideoChatGPT~\cite{videochatgpt} into conversation format based on the video IDs.
\textbf{(2) Simple Caption}
aims to improve basic visual description capabilities.
We choose the widely used COCO Caption~\cite{coco} and WebVid~\cite{bain2021frozen},
together with first-order video captions from YouCook2~\cite{youcook2}.
\textbf{(3) Detailed Caption}
aims at enriching the comprehensive capabilities for understanding visual details.
We leverage the detailed caption data from 
MiniGPT-4~\cite{minigpt4}, 
LLaVA~\cite{llava} and VideoChat~\cite{videochat}.
We also integrate Paragraph Captioning~\cite{paragraphs},
TextCaps~\cite{textcaps},
and TextVR~\cite{textvr},
which require uniquely comprehending text within images and videos.
\textbf{(4) VQA} aims to improve visual question-answering capabilities.
We include the basic VQA (VQAv2~\cite{vqa}, GQA~\cite{gqa}, TGIF-QA~\cite{tgif_qa} and WebVidQA~\cite{just_ask}), 
knowledge-based VQA (OK-VQA~\cite{okvqa}, AOK-VQA~\cite{aokvqa} and ViQuAE~\cite{viquae}), 
OCR-based VQA (OCR-VQA~\cite{ocr_vqa}, TextVQA~\cite{textvqa}, ST-VQA~\cite{st_vqa} and DocVQA~\cite{docvqa}), 
and egocentric VQA from Ego4D~\cite{ego4d}.
\textbf{(5) Reasoning} focuses on enhancing diverse reasoning capacities.
We use 
LLaVA-reasoning~\cite{llava} and CLEVR~\cite{clevr} for spatial reasoning, 
VisualMRC~\cite{visualmrc} for reading comprehension, 
NExT-QA~\cite{nextqa} for temporal reasoning, 
and CLEVRER~\cite{clevrer} for spatiotemporal reasoning.
\textbf{(6) Classification} aims at boosting robustness to object and action recognition.
We sample data from ImageNet~\cite{imagenet}, COCO-ITM~\cite{coco}, Kinetics-710~\cite{uniformerv2} and SthSthV2~\cite{sth}. 

\subsection{Progressive Multi-Modal Training}

Another critical factor in boosting MLLMs is how to effectively bridge the semantic gap between visual and linguistic representation.
To tackle this problem,
we adopt a progressive multi-modal training paradigm as shown in Fig. \ref{fig:model}.

\textbf{Stage1: Vision-Language Alignment.}
In the first stage,
we aim at aligning vision and text.
To balance efficiency and effectiveness,
we freeze the visual encoder and train a flexible QFormer~\cite{blip2},
which compresses redundant visual tokens into fewer query tokens,
and align these queries with text tokens by multi-modal losses,
\textit{i.e.},
Vision-Text Contrastive learning (VTC), 
Vision-Text Matching (VTM), 
and Vision-grounded Text Generation (VTG). 
But different from~\cite{blip2},
we choose the pretrained UMT-L~\cite{umt} as our visual encoder,
due to its powerful capability of spatial-temporal representation learning.
Moreover,
we train QFormer with only 15M image captions from CC3M~\cite{cc3m} and CC12M~\cite{cc12m} but 10M video captions from WebVid-10M~\cite{bain2021frozen},
in order to enhance video-language modeling.

\input{tables/mvpbench}

\input{tables/videochatgpt_benchmark}
\input{tables/zs_vqa}

\textbf{Stage2: Vision-Language Connection.}
After initial alignment,
we then connect the visual encoder with the pretrained LLMs,
for building vision-language understanding capabilities.
Following~\cite{blip2},
we apply a linear projection to further transform the query tokens,
and concatenate the projected tokens with the text tokens into LLM for vision-based caption generation (\textit{i.e.}, VTG).
But different from~\cite{blip2},
we unfreeze the visual encoder for better alignment with LLM.
In addition to the aforementioned training data in Stage1, we further introduce 2M image captions (COCO~\cite{coco}, Visual Genome~\cite{vg}, and SBU~\cite{sbu}) and 10M video captions (InternVid~\cite{internvid}),
to enrich the caption diversity.

\textbf{Stage3: Instruction Tuning.}
In the final stage, 
we employ the proposed data in Section \ref{sec:instruction_data} for instruction tuning. 
To better align responses with instructions,
we use low-rank adaptation~\cite{lora} on the frozen LLM,
and tune it along with the visual encoder and QFormer by VTG loss.
Moreover,
inspired by~\cite{instructblip},
we integrate instructions 
(\textit{i.e.}, `\texttt{i}' of `\texttt{QA}') into QFormer,
in order to extract instruction-relevant visual tokens as input to LLM.
However,
different from~\cite{instructblip},
we do not incorporate questions (\textit{i.e.}, `\texttt{q}' of `\texttt{QA}') into QFormer due to the subpar performances (see appendix.).

%% file: tables/mvpbench.tex
\begin{table*}[tp]
    \centering
    \setlength\tabcolsep{1pt}
    \resizebox{1.0\textwidth}{!}{
        \begin{tabular}{l|l|c|c|c|c|c|c|c|c|c|c|c|c|c|c|c|c|c|c|c|c|c}
        \Xhline{1.0pt}
        \textbf{Model} & \textbf{LLM} & \textbf{Avg} & \textbf{AS} & \textbf{AP} & \textbf{AA} & \textbf{FA} & \textbf{UA} & \textbf{OE} & \textbf{OI} & \textbf{OS} & \textbf{MD} & \textbf{AL} & \textbf{ST} & \textbf{AC} & \textbf{MC} & \textbf{MA} & \textbf{SC} & \textbf{FP} & \textbf{CO} & \textbf{EN} & \textbf{ER} & \textbf{CI} \\
        \Xhline{1.0pt}
        Random & - & \cellcolor{gray!20}{27.3} & 25.0 & 25.0 & 33.3 & 25.0 & 25.0 & 33.3 & 25.0 & 33.3 & 25.0 & 25.0 & 25.0 & 33.3 & 25.0 & 33.3 & 33.3 & 25.0 & 33.3 & 25.0 & 20.0 & 30.9 \\
        \Xhline{0.8pt}
        \multicolumn{23}{l}{\gray{\textit{\textbf{Image MLLMs:} Following~\cite{instructblip}, all models take \textbf{4} frames as input, with the output embeddings concatenated before feeding into the LLM.}}} \\
        mPLUG-Owl-I~\cite{mplug-owl} & LLaMA-7B & \cellcolor{gray!20}{29.4} & 25.0 & 20.0 & 44.5 & 27.0 & 23.5 & 36.0 & 24.0 & 34.0 & 23.0 & 24.0 & 34.5 & 34.5 & 22.0 & 31.5 & 40.0 & 24.0 & 37.0 & 25.5 & 21.0 & 37.0 \\
        LLaMA-Adapter~\cite{llamaadapter} & LLaMA-7B & \cellcolor{gray!20}{31.7} & 23.0 & 28.0 & 51.0 & 30.0 & 33.0 & 53.5 & 32.5 & 33.5 & 25.5 & 21.5 & 30.5 & 29.0 & 22.5 & 41.5 & 39.5 & 25.0 & 31.5 & 22.5 & 28.0 & 32.0 \\
        BLIP2~\cite{blip2} & FlanT5-XL & \cellcolor{gray!20}{31.4} & 24.5 & 29.0 & 33.5 & 17.0 & 42.0 & 51.5 & 26.0 & 31.0 & 25.5 & 26.0 & 32.5 & 25.5 & 30.0 & 40.0 & 42.0 & 27.0 & 30.0 & 26.0 & 37.0 & 31.0 \\
        Otter-I~\cite{otter} & MPT-7B & \cellcolor{gray!20}{33.5} & 34.5 & 32.0 & 39.5 & 30.5 & 38.5 & 48.5 & 44.0 & 29.5 & 19.0 & 25.5 & 55.0 & 20.0 & 32.5 & 28.5 & 39.0 & 28.0 & 27.0 & 32.0 & 29.0 & 36.5 \\
        MiniGPT-4~\cite{minigpt4} & Vicuna-7B & \cellcolor{gray!20}{18.8} & 16.0 & 18.0 & 26.0 & 21.5 & 16.0 & 29.5 & 25.5 & 13.0 & 11.5 & 12.0 & 9.5 & 32.5 & 15.5 & 8.0 & 34.0 & 26.0 & 29.5 & 19.0 & 9.9 & 3.0 \\
        InstructBLIP~\cite{instructblip} & Vicuna-7B & \cellcolor{gray!20}{32.5} & 20.0 & 16.5 & 46.0 & 24.5 & 46.0 & 51.0 & 26.0 & 37.5 & 22.0 & 23.0 & 46.5 & \textbf{42.5} & 26.5 & 40.5 & 32.0 & 25.5 & 30.0 & 25.5 & 30.5 & 38.0 \\
        LLaVA~\cite{llava} & Vicuna-7B & \cellcolor{gray!20}{36.0} & 28.0 & 39.5 & 63.0 & 30.5 & 39.0 & 53.0 & 41.0 & 41.5 & 23.0 & 20.5 & 45.0 & 34.0 & 20.5 & 38.5 & 47.0 & 25.0 & 36.0 & 27.0 & 26.5 & 42.0 \\
        \Xhline{0.8pt}
        \multicolumn{23}{l}{\gray{\textit{\textbf{Video MLLMs:} All models take \textbf{16} frames as input, with the exception of VideoChatGPT, which uses \textbf{100} frames.}}} \\
        Otter-V~\cite{otter} & LLaMA-7B & \cellcolor{gray!20}{26.8} & 23.0 & 23.0 & 27.5 & 27.0 & 29.5 & 53.0 & 28.0 & 33.0 & 24.5 & 23.5 & 27.5 & 26.0 & 28.5 & 18.0 & 38.5 & 22.0 & 22.0 & 23.5 & 19.0 & 19.5 \\
        mPLUG-Owl-V~\cite{mplug-owl} & LLaMA-7B & \cellcolor{gray!20}{29.7} & 22.0 & 28.0 & 34.0 & 29.0 & 29.0 & 40.5 & 27.0 & 31.5 & \textbf{27.0} & 23.0 & 29.0 & 31.5 & 27.0 & 40.0 & 44.0 & 24.0 & 31.0 & 26.0 & 20.5 & 29.5 \\
        VideoChatGPT~\cite{videochatgpt} & Vicuna-7B & \cellcolor{gray!20}{32.7} & 23.5 & 26.0 & 62.0 & 22.5 & 26.5 & 54.0 & 28.0 & 40.0 & 23.0 & 20.0 & 31.0 & 30.5 & 25.5 & 39.5 & \textbf{48.5} & 29.0 & 33.0 & 29.5 & 26.0 & 35.5 \\
        VideoLLaMA~\cite{videollama} & Vicuna-7B & \cellcolor{gray!20}{34.1} & 27.5 & 25.5 & 51.0 & 29.0 & 39.0 & 48.0 & 40.5 & 38.0 & 22.5 & 22.5 & 43.0 & 34.0 & 22.5 & 32.5 & 45.5 & 32.5 & 40.0 & 30.0 & 21.0 & 37.0 \\
        VideoChat~\cite{videochat} & Vicuna-7B & \cellcolor{gray!20}{35.5} & 33.5 & 26.5 & 56.0 & 33.5 & 40.5 & 53.0 & 40.5 & 30.0 & 25.5 & 27.0 & 48.5 & 35.0 & 20.5 & 42.5 & 46.0 & 26.5 & 41.0 & 23.5 & 23.5 & 36.0 \\
        \hline
        \textbf{\ModelName$_\mathbf{text}$} & Vicuna-7B & \cellcolor{gray!20}{34.7} & 24.5 & 27.0 & 49.5 & 27.0 & 38.0 & 53.0 & 28.0 & 40.0 & 25.5 & 27.0 & 38.5 & 41.5 & 27.5 & 32.5 & 46.5 & 26.5 & 36.0 & 33.0 & 32.0 & 40.0 \\
        \textbf{\ModelName} & Vicuna-7B & \cellcolor{gray!20}{\textbf{51.1}} & \textbf{66.0} & 47.5 & \textbf{83.5} & \textbf{49.5} & 60.0 & \textbf{58.0} & \textbf{71.5} & \textbf{42.5} & 23.0 & 23.0 & \textbf{88.5} & 39.0 & \textbf{42.0} & \textbf{58.5} & 44.0 & \textbf{49.0} & 36.5 & \textbf{35.0} & 40.5 & \textbf{65.5} \\
        \Xhline{1.0pt}
        \multicolumn{23}{l}{\gray{\textit{GPT-4V take \textbf{16} frames as input, and the resolution is \textbf{512$\times$512}, while others use small resolution of \textbf{224$\times$224}.}}} \\
        GPT-4V~\cite{gpt4v} & GPT-4 & \cellcolor{gray!20}{43.5} & 55.5 & \textbf{63.5} & 72.0 & 46.5 & \textbf{73.5} & 18.5 & 59.0 & 29.5 & 12.0 & 40.5 & \textbf{83.5} & \textbf{39.0} & 12.0 & 22.5 & 45.0 & 47.5 & 52.0 & 31.0 & \textbf{59.0} & 11.0 \\
        \textbf{\ModelName} & Mistral-7B & \cellcolor{gray!20}{\textbf{60.4}} & \textbf{75.5} & 58.0 & \textbf{83.5} & \textbf{50.5} & 60.5 & \textbf{87.5} & \textbf{74.5} & \textbf{45.0} & \textbf{47.5} & \textbf{44.0} & 82.5 & 37.0 & \textbf{64.5} & \textbf{87.5} & \textbf{51.0} & \textbf{66.5} & \textbf{47.0} & \textbf{35.0} & 37.0 & \textbf{72.5} \\
        \Xhline{1.0pt}
        \end{tabular}
    }
    \vspace{-0.3cm}
    \caption{
    \textbf{Evaluations results on \BenchName.}
    Excluding BLIP2 and Otter, 
    all models are built upon \textbf{LLaMA 1}~\cite{llama1}
    for fair comparisons by default. 
    ``\textbf{Random}'' refers to results from random guesses.
    ``\textbf{\ModelName$_\mathbf{text}$}'' denotes the model receiving blank videos and excludes LoRA tuning, relying solely on the LLM's capacity for responses.
    Full results on MVBench can be found at \url{https://huggingface.co/spaces/OpenGVLab/MVBench_Leaderboard}.
    Notably, \textbf{our \ModelName\  exceeds the leading models by over 15\%.
    Built upon Mistral~\cite{mistral},
    our \ModelName\  significantly outperforms GPT-4V~\cite{gpt4v} by 16.9\%.}}
    \label{tab:mvpbench}
    \vspace{-0.3cm}
\end{table*}

%% file: tables/videochatgpt_benchmark.tex
\begin{table}[tp]
    \centering
    \vspace{-0.2cm}
    \setlength\tabcolsep{2.0pt}
    \resizebox{1.0\linewidth}{!}{
        \begin{tabular}{l|c|c|c}
        \Xhline{1.0pt}
        \textbf{Evaluation Aspect} &
        \small{VideoChat\tiny{\cite{videochat}}} & \small{VideoChatGPT\tiny{\cite{videochatgpt}}} & \textbf{\small{\ModelName}} \\
        \Xhline{0.8pt}
        \small{Correctness of Information} & 2.23 & 2.40 & \textbf{3.02} \\ 
        \small{Detail Orientation} & 2.50 & 2.52 & \textbf{2.88} \\
        \small{Contextual Understanding} & 2.53 & 2.62 & \textbf{3.51} \\
        \small{Temporal Understanding} & 1.94 & 1.98 & \textbf{2.66} \\ 
        \small{Consistency} & 2.24 & 2.37 & \textbf{2.81} \\
        \hline
        \textbf{Avg} & 2.29 & 2.38 & \textbf{2.98} \\
        \Xhline{1.0pt}
        \end{tabular}
    }
    \vspace{-0.3cm}
    \caption{\textbf{Results of video conversation benchmark~\cite{videochatgpt}.}}
    \label{tab:videochatgpt_benchmark}
    \vspace{-0.4cm}
\end{table}

%% file: tables/zs_vqa.tex
\begin{table}[tp]
    \centering
    \setlength\tabcolsep{3.2pt}
    \resizebox{1.0\linewidth}{!}{
        \begin{tabular}{l|cc|cc|cc}
        \Xhline{1.0pt}
        \multirow{2}{*}{\textbf{Model}}  & \multicolumn{2}{c|}{\textbf{MSVD-QA}} & \multicolumn{2}{c|}{\textbf{MSRVTT-QA}} & \multicolumn{2}{c}{\textbf{ANet-QA}} \\
        ~ & \textbf{Acc} & \textbf{Score} & \textbf{Acc} & \textbf{Score} & \textbf{Acc} & \textbf{Score} \\ 
        \hline
        VideoLLaMA~\cite{videollama}  & 51.6 & 2.5 & 29.6 & 1.8 & 12.4 & 1.1 \\ 
        VideoChat~\cite{videochat}  & 56.3 & 2.8 & 45.0 & 2.5 & 26.5 & 2.2 \\
        VideoChatGPT~\cite{videochatgpt}  & 64.9 & 3.3 & 49.3 & 2.8 & 35.2 & 2.7 \\ 
        \hline
        \textbf{\ModelName} & \textbf{70.0} & \textbf{3.9} & \textbf{54.1} & \textbf{3.3} & \textbf{49.1} & \textbf{3.3} \\
        \Xhline{1.0pt}
        \end{tabular}
    }
    \vspace{-0.3cm}
    \caption{\textbf{Zero-shot video QA results on~\cite{msrvtt_qa,anet_qa}.}}
    \label{tab:zs_vqa}
    \vspace{-0.3cm}
\end{table}

%% file: sec/4_experiments.tex
\section{Experiments}
\label{sec:exp}

\textbf{Implementation Details}.
For visual encoder and LLM,
we apply UMT-L~\cite{umt} and Vicuna-7B v0~\cite{vicuna} by default.
Following BLIP2~\cite{blip2},
we deploy QFormer using the pretrained BERT$_{base}$~\cite{devlin2018bert}.
32 queries are used in Stage1,
and 
extra 64 queries are introduced in Stage2 and Stage3 when the visual encoder is unfrozen.
For efficient training, 
4-frame videos are processed through 10 epochs in Stage1 and 1 epoch in Stage2. 
Transitioning to Stage3, 
we shift to 8-frame videos for 3 epochs.
For evaluation,
we input 16-frame videos with elaborate prompts for better results.

\subsection{Results on \BenchName}
Tab.~\ref{tab:mvpbench} presents the evaluation results on \BenchName, 
revealing that current image and video MLLMs are underperforming.
For instance, 
VideoChat~\cite{videochat}, a top-performing video MLLM, only marginally surpasses \textbf{\ModelName$_\mathbf{text}$} by 0.8\% in average accuracy (35.5\% \textit{vs.} 34.7\%), 
with the latter generating responses from text alone. 
In contrast, our \textbf{\ModelName}\  markedly exceeds the leading model by over \textbf{15\%}, 
particularly excelling in categories like action, object, scene, attribute, and pose recognition.
However, 
it struggles in position, count, and character tasks,
performing less effectively than \ModelName$_\mathbf{text}$,
which could be attributed to the lack of exposure to these tasks during instruction tuning.

Surprisingly,
built upon Mistral~\cite{mistral} with SMiT~\cite{smit} instructions,
our \ModelName\ significantly improves the results,
delivering strong performances across various tasks. 

Furthermore, we evaluated the powerful GPT-4V~\cite{gpt4v}. 
The results show that while GPT-4V achieves satisfactory performance, 
demonstrating its capacity for temporal understanding, 
our \ModelName\ surpasses it, 
increasing accuracy by 16.9\%. 
This further underscores the superiority of our model in handling a broader range of tasks.

More results on NExT-QA~\cite{nextqa}, STAR~\cite{star}, TVQA~\cite{tvqa}, EgoSchema~\cite{egoschema} and IntentQA~\cite{intentqa} can be found in the appendix.
Our \ModelName\ demonstrates robust performance on these complex reasoning tasks.

\begin{figure}[t]
    \centering
    \vspace{-0.2cm}
    \includegraphics[width=0.95\linewidth
    ]{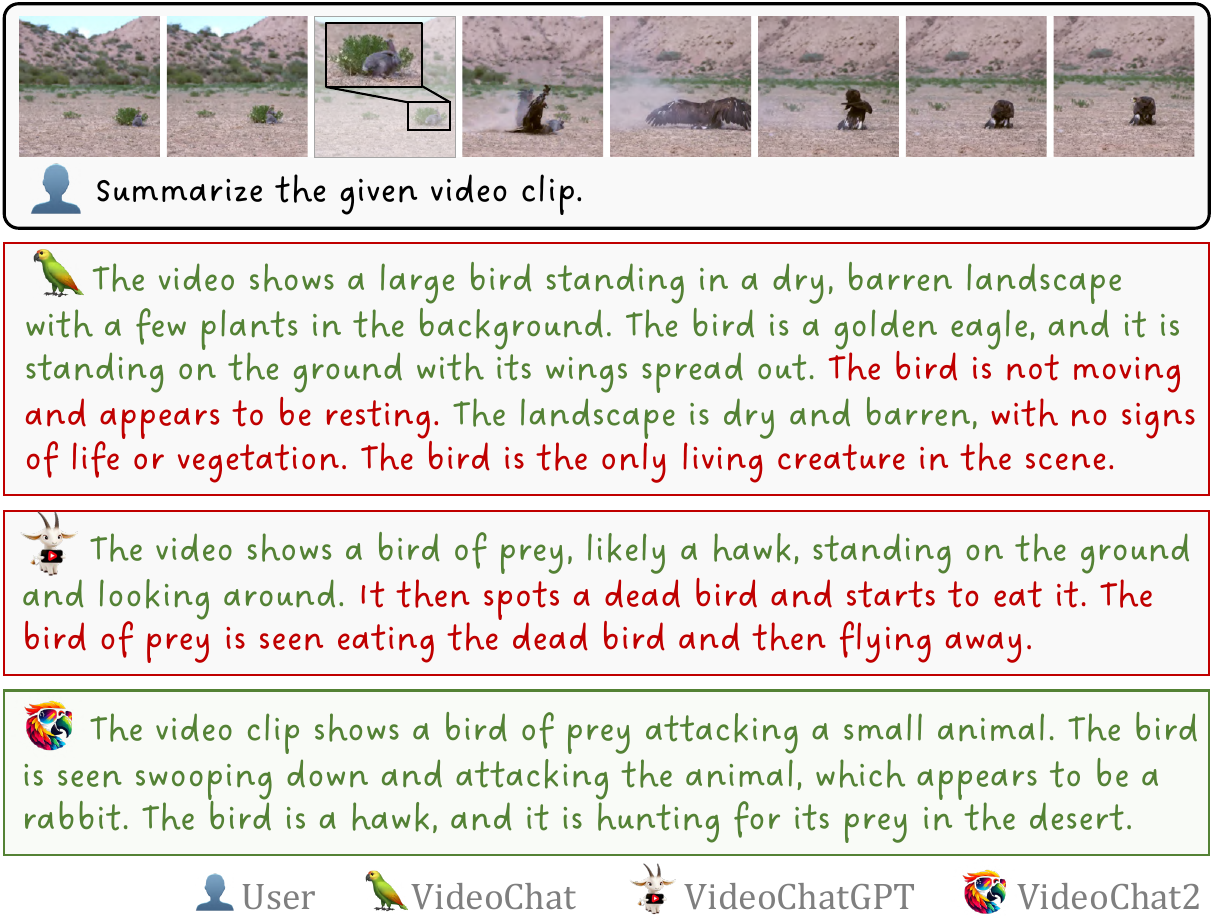}
    \vspace{-0.35cm}
    \caption{
    \textbf{Qualitative comparison.}
    \darkGreen{Green} signifies accurate descriptions, while \red{red} denotes incorrect or hallucinatory responses.
    }
    \label{fig:example1}
    \vspace{-0.3cm}
\end{figure}

\subsection{More Comparisons}
Following~\cite{videochatgpt},
we use ChatGPT~\cite{chatgpt} to conduct quantitative comparisons among video MLLMs.
\textbf{(1) Video Conversation:}
Tab. \ref{tab:videochatgpt_benchmark} shows the results on the benchmark of \cite{videochatgpt}.
Compared with VideoChatGPT~\cite{videochatgpt},
our \ModelName\  exhibits superior performances across all metrics,
with distinct advancements in terms of information correctness as well as context and temporal understanding.
This indicates that our \ModelName\ is more adept at comprehending both spatial and temporal details
and providing consistent and reliable responses.
\textbf{(2) Zero-Shot Video QA:}
Tab. \ref{tab:zs_vqa} lists results of typical video QA datasets~\cite{msrvtt_qa,activitynet_qa}.
It is evident that our \ModelName\  surpasses all other methods,
particularly excelling in understanding long videos in ActivityNet~\cite{activitynet_qa}.

We further present a qualitative comparison in Fig. \ref{fig:example1},
where \ModelName\ delivers a precise and thorough response. 
For more qualitative analyses,
see the appendix.

\input{tables/ablation_data}

\input{tables/ablation_llm}

\input{tables/ablation_stage}

\subsection{Ablations of \ModelName}
\label{sec:ablation}
In this section, we conduct comprehensive analyses of the instruction data, model architecture, and prompt designs.

\textbf{Instruction Data.}
Tab. \ref{tab:ablation_data} demonstrates that the limited instruction data proposed in VideoChat~\cite{videochat} (17K) and VideoChatGPT~\cite{videochatgpt} (100K) is insufficient for temporal understanding. 
As we increase the data diversity and quantity, 
the performances are significantly improved, 
wherein video data contributes more than image data (50.5\% \textit{vs.} 42.1\%). 
Considering the potential redundancy in the simple caption data of COCO~\cite{coco} and WebVid~\cite{bain2021frozen}, 
we randomly compress them. 
This results in only a minimal impact on performance (50.7\% \textit{vs.} 51.1\%), 
while accelerating the tuning by 1.7$\times$.

\textbf{Architecture.}
\textbf{(1) Visual Encoder:}
In Tab. \ref{tab:ablation_llm}, we first apply EVA-CLIP-g~\cite{eva_clip} akin to VideoChat, 
which achieves 6.9\% higher accuracy with our instruction data (42.4\% \textit{vs.} 35.5\% for original one in Tab. \ref{tab:mvpbench}).
Further substitutions with UMT-L improve the performance by an additional 6.2\%,
which demonstrates the effectiveness of our visual encoder.
\textbf{(2) LLM:}
However,
incorporating larger and newer LLMs offers a marginal improvement in the results, 
indicating that \BenchName\  relies predominantly on the visual encoder. 
Notably,
LoRA~\cite{lora} consistently uplifts the results, 
potentially due to its enhanced capacity for instruction following.

\input{tables/ablation_system_prompt}

\input{tables/ablation_answer_prompt}

\textbf{Training Method.}
Initially, 
we tune only the linear projection while freezing the visual encoder and QFormer as in MiniGPT-4~\cite{minigpt4},
but it yielded subpar results in Tab. \ref{tab:ablation_training}. 
By unfreezing QFormer as ~\cite{instructblip}, 
we achieve an 8.5\% performance boost. 
Further, when we unfreeze the visual encoder, 
results consistently improved, 
emphasizing the value of more learnable parameters for visual adaptation.

\textbf{Prompt Design.}
Tab. \ref{tab:ablation_system_prompt} reveals that a comprehensive \textit{system prompt}, which underscores the task requirement, enhances task completion effectiveness.
Different from the unstable ChatGPT-extracting methods~\cite{mmbench} and more time-consuming log-likelihood comparisons~\cite{seedbench}, 
we apply a simple yet effective \textit{answer prompt} to extra the options.
Results in Tab. \ref{tab:ablation_answer_prompt} demonstrate that it accurately targets the option and enhances response precision across various MLLMs.
More importantly,
\ModelName\ follows the instructions better to return options even without the prompt.

%% file: tables/ablation_data.tex
\begin{table}[tp]
    \centering
    \setlength\tabcolsep{4.0pt}
    \resizebox{1.0\linewidth}{!}{
        \begin{tabular}{l|l|l|l|l}
        \Xhline{1.0pt}
        \textbf{Data Source} & \multicolumn{1}{c|}{\textbf{Type}} & \multicolumn{1}{c|}{\textbf{Task}} & \multicolumn{1}{c|}{\textbf{\#Num}} & \multicolumn{1}{c}{\textbf{Avg}} \\
        \Xhline{0.8pt}
        VideoChat~\cite{videochat}
        & I$+$V & DC$+$R$+$C & 17K & 36.4 \\
        \hline
        VideoChatGPT~\cite{videochatgpt} & V & DC & 100K & 34.3 \red{$\downarrow$2.1}  \\
        \hline
        \multirow{4}{*}{\textbf{Ours}} & I & ALL & 1.1M & 42.1 \darkGreen{$\uparrow$5.7} \\
        ~ & V & ALL & 0.9M & 50.5 \darkGreen{$\uparrow$14.1}\\
        ~ & I$+$V\blue{$\dag$} & ALL & 1.2M  & 50.7 \darkGreen{$\uparrow$14.3}  \\
        ~ & \cellcolor{gray!20}{I$+$V} & \cellcolor{gray!20}{ALL}  & \cellcolor{gray!20}{2.0M} & \cellcolor{gray!20}{\textbf{51.1} \darkGreen{$\uparrow$\textbf{14.7}}} \\
        \Xhline{1.0pt}
        \end{tabular}
    }
    \vspace{-0.3cm}
    \caption{
    \textbf{Instruction Data.}
    ``I'' and ``V'' denote ``Image'' and ``Video'',
    while ``DC'', ``R'', ``C'' represent ``Detailed Caption'', ``Reasoning'' and ``Conversation''.
    ``\blue{$\dag$}'' symbolizes the version with fewer captions: 100K from COCO~\cite{coco}, 80K from WebVid~\cite{bain2021frozen}.
    }
    \label{tab:ablation_data}
    \vspace{-0.4cm}
\end{table}

%% file: tables/ablation_llm.tex
\begin{table}[tp]
    \centering
    \setlength\tabcolsep{6.0pt}
    \resizebox{1.0\linewidth}{!}{
        \begin{tabular}{l|l|c|l}
        \Xhline{1.0pt}
        \textbf{Visual Encoder} &
        \textbf{LLM} & \textbf{LoRA} & \multicolumn{1}{c}{\textbf{Avg}} \\
        \Xhline{0.8pt}
        \multirow{2}{*}{EVA-CLIP-g~\cite{eva_clip}} & \multirow{2}{*}{Vicuna-7B \textit{v0}} & \xmark & 42.4 \\
        ~ & ~ & \cmark & 45.3 \darkGreen{$\uparrow$2.9} \ \\
        \hline
        \multirow{6}{*}{\textbf{UMT-L~\cite{umt}}} & \multirow{2}{*}{Vicuna-7B \textit{v0}} & \xmark & 48.6 \\
        ~ & ~ & \cmark & 51.1 \darkGreen{$\uparrow$2.5}\ \\
        \cline{2-4}
        ~ & Vicuna-13B \textit{v0} & \cmark & 51.4 \\
        \cline{2-4}
        ~ & \multirow{2}{*}{Vicuna-7B \textit{v1.5}} & \xmark & 48.1 \\
        ~ & ~ & \cmark & 51.2 \darkGreen{$\uparrow$3.1} \\
        \cline{2-4}
        ~ & Vicuna-13B \textit{v1.5} & \cmark & 51.6 \\
        \Xhline{1.0pt}
        \end{tabular}
    }
    \vspace{-0.3cm}
    \caption{\textbf{Visual Encoder \& LLM.}
    Vicuna~\cite{vicuna} \textit{v0} and \textit{v1.5} models are tuned from LLaMA 1~\cite{llama1} and LLaMA 2~\cite{llama2} respectively.
    }
    \label{tab:ablation_llm}
    \vspace{-0.4cm}
\end{table}

%% file: tables/ablation_stage.tex
\begin{table}[tp]
    \centering
    \setlength\tabcolsep{2.5pt}
    \resizebox{1.0\linewidth}{!}{
        \begin{tabular}{c|c|c|c|l}
        \Xhline{1.0pt}
        \multicolumn{2}{c|}{\textbf{Stage2}} & \multicolumn{2}{c|}{\textbf{Stage3}} & \multicolumn{1}{c}{\multirow{2}{*}{\textbf{Avg}}} \\
        \textbf{\small{Visual Encoder}} & \textbf{\small{QFomer}} & \textbf{\small{Visual Encoder}} & \textbf{\small{QFomer}} & ~ \\
        \Xhline{0.8pt}
        \ice & \ice & \ice & \ice & 38.5 \\
        \hline
        \ice & \fire & \ice & \fire & 47.0 \darkGreen{$\uparrow$8.5}  \\
        \hline
        \fire & \fire & \ice & \fire & 47.5 \darkGreen{$\uparrow$9.0}  \\
        \hline
        \rowcolor{gray!20} 
        \fire & \fire & \fire & \fire & \textbf{51.1} \darkGreen{$\uparrow$\textbf{12.6}}  \\
        \Xhline{1.0pt}
        \end{tabular}
    }
    \vspace{-0.3cm}
    \caption{\textbf{Training Method.} \ice and\fire refer to freezing and tuning.
    We efficiently freeze the visual encoder in Stage1 and LLM in all stages,
    while tuning the visual encoder and QFormer in Stage2\&3.
    }
    \label{tab:ablation_training}
    \vspace{-0.3cm}
\end{table}

%% file: tables/ablation_system_prompt.tex
\begin{table}[tp]
    \centering
    \setlength\tabcolsep{2.0pt}
    \resizebox{1.0\linewidth}{!}{
        \begin{tabular}{l|c}
        \Xhline{1.0pt}
        \textbf{System Prompt} & \textbf{Avg} \\
        \Xhline{0.8pt}
        \textit{\makecell[l]{Carefully observe the video and choose the best option\\ for the question.}} & 49.9 \\
        \hline
        \textit{\makecell[l]{Carefully watch the video and \darkGreen{pay attention to the cause,} \\ \darkGreen{sequence of events, and object details and movements.} \\Based on your observations, select the best option that\\ accurately addresses the question.}} & \makecell[c]{50.5\\\darkGreen{$\uparrow$0.6}} \\
        \hline
        \rowcolor{gray!20} 
        \textit{\makecell[l]{Carefully watch the video and \darkGreen{pay attention to the cause} \\\darkGreen{and sequence of events, the detail and movement of} \\\darkGreen{objects and the action and pose of persons.} \\Based on your observations, select the best option that \\accurately addresses the question.}} & \textbf{\makecell[c]{51.1\\\darkGreen{$\uparrow$1.2}}}\\
        \Xhline{1.0pt}
        \end{tabular}
    }
    \vspace{-0.3cm}
    \caption{\textbf{System Prompt.} It should consider temporal evolution.}
    \label{tab:ablation_system_prompt}
    \vspace{-0.4cm}
\end{table}

%% file: tables/ablation_answer_prompt.tex
\begin{table}[tp]
    \centering
    \setlength\tabcolsep{3.0pt}
    \resizebox{1.0\linewidth}{!}{
        \begin{tabular}{l|l|c|l}
        \Xhline{1.0pt}
        \textbf{Model} &
        \textbf{Answer Prompt} & \textbf{Hit Ratio} & \multicolumn{1}{c}{\textbf{Avg}} \\
        \Xhline{0.8pt}
        \multirow{2}{*}{VideoChat~\cite{videochat}} & $\varnothing$ & 78.2\% & 22.8 \\
        ~ & \textit{Best option: (} & 100\% & 35.5 \darkGreen{$\uparrow$12.7} \\
        \hline
        \multirow{2}{*}{VideoChatGPT~\cite{videochatgpt}} & $\varnothing$ & 64.6\% & 22.0 \\
        ~ & \textit{Best option: (} & 100\% & 32.8 \darkGreen{$\uparrow$10.8} \\
        \hline
        \multirow{2}{*}{\textbf{\ModelName}} & $\varnothing$ & 96.4\% & 50.1 \\
        ~ & \cellcolor{gray!20}{\textit{Best option: (}} & \cellcolor{gray!20}{100\%} & \cellcolor{gray!20}{51.1 \darkGreen{$\uparrow$1.0}} \\
        \Xhline{1.0pt}
        \end{tabular}
    }
    \vspace{-0.3cm}
    \caption{\textbf{Answer Prompt.} 
    `$\varnothing$' indicates directly matching the option within responses, similar to~\cite{mmbench}.
    Our simple yet effective prompt enhances response precision across various MLLMs.
    }
    \label{tab:ablation_answer_prompt}
    \vspace{-0.3cm}
\end{table}

%% file: sec/5_conclusion.tex
\section{Conclusion}
\label{sec:conclusion}

This paper introduces \BenchName, 
a comprehensive benchmark for evaluating the temporal understanding capabilities of MLLMs. 
Moreover,
we propose a robust video MLLM baseline,
\ModelName,
outperforming the leading models by over 15\% on \BenchName.
Our extensive analyses further direct the designs of MLLMs for temporal understanding.

%% file: sec/6_acknowledgement.tex
\subsection*{Acknowledgement}
This work was supported in part by the National Key R\&D Program of China (No. 2022ZD0160505), and the National Natural Science Foundation of China under Grant (62272450, 62076119).

%% file: sec/X_suppl.tex
\appendix

\section{Training Hyperparameters}
The hyperparameters used in different stages of training are listed in Tab.~\ref{tab:hyperparameters}. 
We adopt TSN~\cite{tsn} sampling for all the videos as previous methods~\cite{umt,videomaev2,videochat}.
For both Stage1 and Stage2, we employ large-scale image and video caption data, as outlined in the main manuscript.
During Stage3, we make use of diverse instruction data and incorporate LoRA modules~\cite{lora} into the LLM with a rank of 16, an alpha value of 32, and a dropout rate of 0.1. 
We apply flash attention~\cite{flashattention} to expedite the training process.

\input{tables/hyperparameters}

\section{More Ablations}
We have carried out further ablation studies, the results of which are displayed in Tabs. \ref{tab:ablation_qformer}, \ref{tab:ablation_data2}, \ref{tab:ablation_resolution}, and \ref{tab:ablation_question_prompt}.

\textbf{QFormer.}
Considering the richer information of video,
we further introduce extra random-initialized queries after Stage1.
Tab. \ref{tab:ablation_qformer} shows that more queries in Stage2 and Stage3 is beneficial, 
leading us to adopt 64 queries by default.
Furthermore, inserting instructions without a question effectively steers toward more accurate responses.
We argue that overly long context (``\textit{instruction} $+$ \textit{question}'') may be difficult for information extraction of QFormer.

\input{tables/ablation_qformer}

\textbf{Resolution \& Frame.}
Tab.~\ref{tab:ablation_resolution} reveals that increasing resolution does not improve performance; however, augmenting the number of frames enhances outcomes. 
This suggests that our \BenchName\  primarily relies on temporal understanding instead of spatial understanding capacity.

\input{tables/ablation_resolution}

\textbf{Instruction data.}
Note that there is a minimal source gap between our instruction data and \BenchName.
Specifically,
the CLEVRER~\cite{clevrer} in our instruction data has similar questions as \textit{Moving Attribute} and \textit{Counterfactual Inference} in \BenchName,
leading the evaluation is not strictly out-domain.
And the videos of \textit{Action Antonym} are from SthSthV2~\cite{sth},
while the antonym is from PAXION~\cite{paxion}.
We try to remove CLEVRER and SthSthV2 in the instruction data to evaluate their impact. 
The results outlined in Tab. \ref{tab:ablation_data2} suggest a more pronounced influence from CLEVRER data, while SthSthV2 data appears to have less effect.

\input{tables/ablation_data2}

\textbf{Question prompt.}
During our experiments, 
we observed that various MLLMs often provide options along with detailed explanations. 
To circumvent this, 
we intentionally craft our question prompts to prevent such detailed outputs. 
Additionally, 
drawing inspiration from the Chain-of-Thought~\cite{cot}, 
we introduce the phrase ``Let's think step by step'' into our prompts to direct the MLLMs' reasoning process. 
However, as indicated by the results in Tab. \ref{tab:ablation_question_prompt}, 
these tactics appear to have negative consequences.

\input{tables/ablation_question_prompt}

\section{Details of QA Generation}

In Tab.~\ref{tab:ben_details}, we present a detailed description of our data generation methodology for \BenchName. 
We have designed various strategies based on different data to increase task difficulty and enhance data diversity.
For those datasets requiring question generation, 
we utilize ChatGPT~\cite{chatgpt} to generate 3 to 5 questions based on the task definitions.

\section{Results on Challenging Video QA}

In Tabs. \ref{tab:ft_nextqa}, \ref{tab:zs_star_tvqa}, \ref{tab:zs_egoschema} and \ref{tab:ft_intentqa}, 
we extend the evaluation of our \ModelName\ to other challenging video benchmarks
\textit{i.e.},
NExT-QA~\cite{nextqa}, STAR~\cite{star}, TVQA~\cite{tvqa}, EgoSchema~\cite{egoschema} and IntentQA~\cite{intentqa}.
Different from the previous methods~\cite{sevila},
which provide answers by comparing the likelihood of different options,
we output the options directly, 
following the protocol of \BenchName.
Our results indicate that \ModelName\ holds its own against current SOTA methods~\cite{Wang2022InternVideoGV,sevila,mplug-owl} on these complex reasoning tasks,
which underscores the effectiveness and robustness of \ModelName,
especially for long videos.

\input{tables/nextqa}
\input{tables/star_tvqa}
\input{tables/egoschema}
\input{tables/intentqa}


\section{Leaderboards and Analyses}
To facilitate a clear comparison of different open-sourced MLLMs,
we present the leaderboards for different tasks on \BenchName\ in Tab.~\ref{tab:leaderboard} (until \red{2023/11/28}).
Overall,
our \ModelName\ achieves the highest rank across 15 tasks.

\textbf{Action \& Pose.}
For tasks associated with action and pose $\mathrm{(a)(b)(c)(d)(e)(p)}$,
our \ModelName\ and VideoChat~\cite{videochat} tends to outperform VideoChatGPT~\cite{videochatgpt},
underscoring the significance of elaborate video backbones~\cite{uniformerv2,umt} for effective action and pose recognition.

\textbf{Object \& Attribute.}
In object-related tasks $\mathrm{(f)(g)(h)}$,
the performance of image MLLM,
\textit{i.e.} LLaVA~\cite{llava},
compares favorably with our \ModelName.
It could be attributed to its potent attribute recognition capabilities, as illustrated in $\mathrm{(n)}$.
Note that VideoChatGPT~\cite{videochatgpt} is tuned from LLaVA, 
thus achieving similar results on these tasks.

\textbf{Position \& Count \& Character.}
In position-related tasks $\mathrm{(i)(j)}$,
none of the models achieve satisfactory results, their performances being analogous to random guessing. 
For counting and character-related tasks $\mathrm{(l)(q)}$,
our \ModelName\  performs similarly and even worse than \ModelName$_\mathrm{text}$\ without videos (as in Tab. \red{2}).
We hypothesize that current MLLMs have difficulty generalizing to localization and counting tasks in the absence of related tuning data. 
Some recent studies~\cite{shikara,Bai2023QwenVLAF,Chen2023MiniGPTv2LL} incorporate grounding data and tune the LLM to enhance localizing and discriminating abilities. 
In our future work,
we will explore improvements in \ModelName's grounding ability.

\input{tables/bench_details}

\input{tables/leaderboard}

\textbf{Scene.}
As presented in Tab. \ref{tab:leaderboard}$\mathrm{(k)}$,
our \ModelName\ excels at scene transition tasks, 
significantly outperforming other models. 
This showcases its sensitivity to background changes, 
making it effective in recognizing camera movements as shown in Fig. \ref{fig:example3}.

\begin{figure*}[t]
    \centering
    \includegraphics[width=0.92\linewidth
    ]{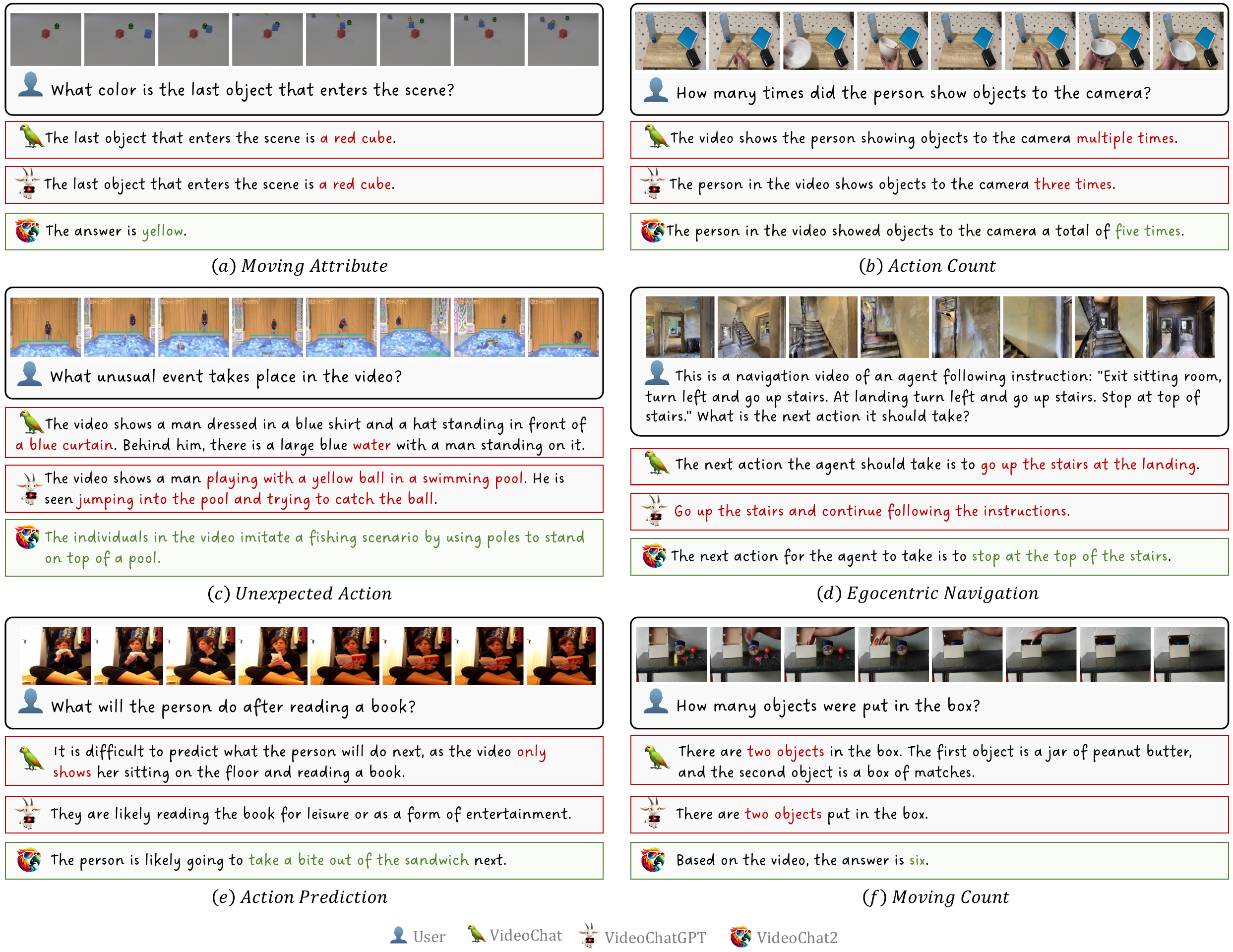}
    \vspace{-0.2cm}
    \caption{\textbf{More qualitative comparisons in \BenchName\ data.} \ModelName\ handles different tasks well.}
    \label{fig:example2}
    \vspace{-0.1cm}
\end{figure*}

\textbf{Cognition.}
In cognition tasks $\mathrm{(r)(s)(t)}$,
our \ModelName\ encounters difficulties with complex egocentric navigation and episode reasoning.
Given the results from FrozenBiLM~\cite{frozenbilm}, 
where the performance for TVQA reasoning significantly improves with the incorporation of speech subtitles, 
we suggest that visual information alone may not be sufficient. The inclusion of other modalities, such as depth and audio, could prove beneficial.

\section{Qualitative Results}
\begin{figure*}[t]
    \centering
    \includegraphics[width=0.92\linewidth
    ]{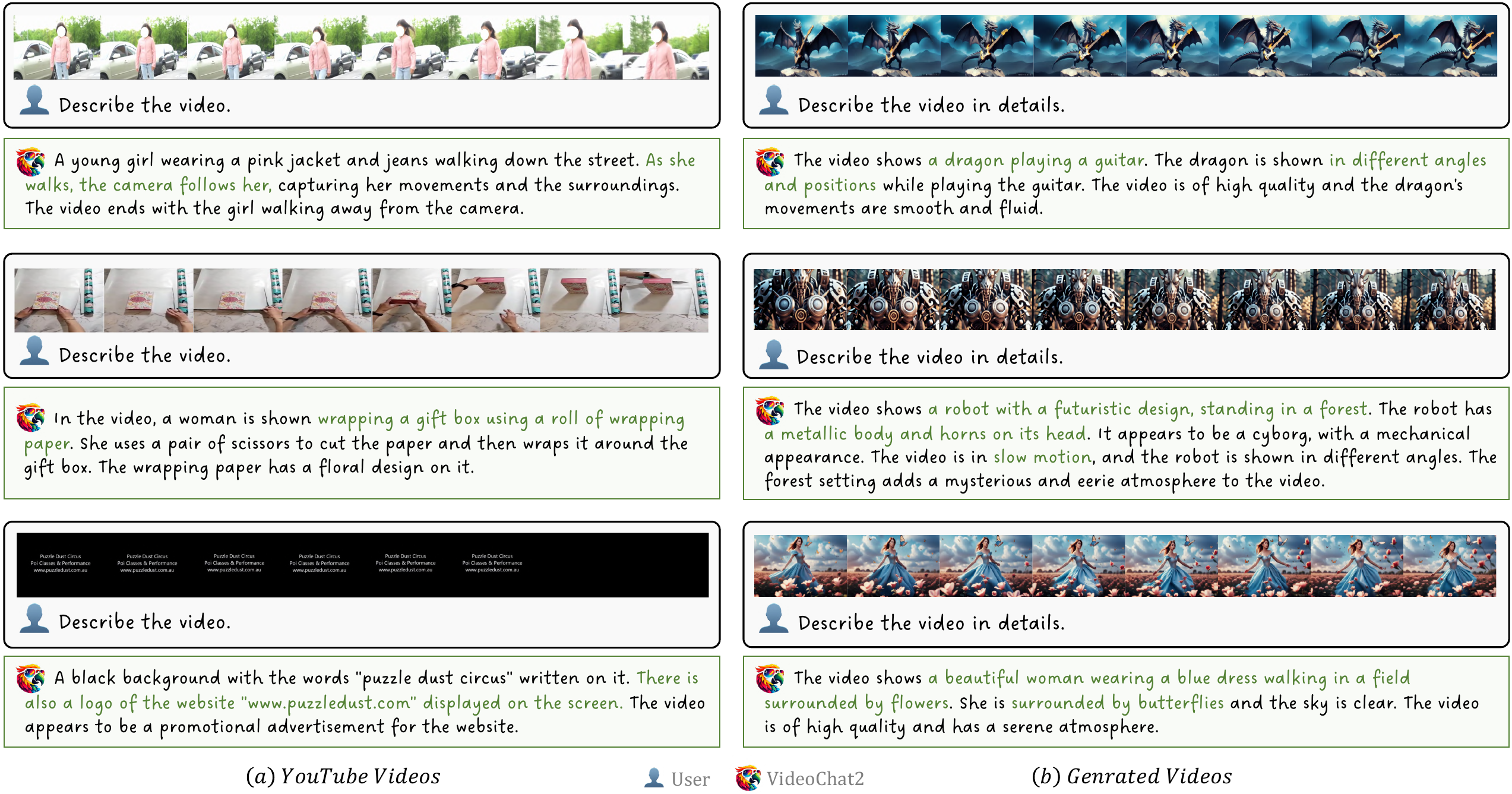}
    \vspace{-0.2cm}
    \caption{\textbf{More descriptive examples.} \ModelName\ can accurately describe the details of diverse videos.}
    \label{fig:example3}
    \vspace{-0.3cm}
\end{figure*}

Additional qualitative results can be found in Figs. \ref{fig:example2} and \ref{fig:example3}. 
Compared with VideoChat~\cite{videochat} and VideoChatGPT~\cite{videochatgpt}, our \ModelName\ performs admirably across a range of tasks in \BenchName.
It possesses the capacity to accurately identify the properties of moving objects, recognize unforeseen actions, and predict future movements based on video context. 
Moreover, it exhibits robustness when dealing with both real and generated videos, adeptly providing detailed insights into human actions, camera motions, background ambiance, and character attributes.


%% file: tables/hyperparameters.tex
\begin{table}[h]
    \centering
    \vspace{-0.1cm}
    \setlength\tabcolsep{10pt}
    \resizebox{1.0\linewidth}{!}{
        \begin{tabular}{l|ccc}
        config & Stage1 & Stage2 & Stage3 \\
        \Xhline{1.0pt}
        input frame & 4 & 4 & 8 \\
        input resolution & 224 & 224 & 224 \\
        max text length & 32 & 32 & 512 \\
        optimizer & \multicolumn{3}{c}{AdamW} \\ 
        optimizer momentum & \multicolumn{3}{c}{$\beta_1, \beta_2{=}0.9, 0.999$}  \\
        weight decay & \multicolumn{3}{c}{0.02} \\
        learning rate schedule & \multicolumn{3}{c}{cosine decay} \\
        learning rate & 1e-4 & 1e-4 & 2e-5 \\
        batch size & 2048 & 512 & 128 \\
        warmup epochs & 1 & 0.2 & 0.6 \\
        total epochs &  10 & 1 & 3 \\
        backbone drop path & \multicolumn{3}{c}{0} \\
        QFormer drop path & \multicolumn{3}{c}{0.2} \\
        QFormer dropout & 0 & 0.1 & 0.1 \\
        QFormer token & 32 & 96 & 96 \\
        flip augmentation &  \multicolumn{3}{c}{\textit{yes}} \\
        augmentation & \multicolumn{3}{c}{MultiScaleCrop [0.5, 1]} \\
        \end{tabular}
    }
    \vspace{-0.3cm}
    \caption{
        \textbf{Training Hyperparameters for different stages.}
    }
    \label{tab:hyperparameters} 
\end{table}

%% file: tables/ablation_qformer.tex
\begin{table}[h]
    \centering
    \setlength\tabcolsep{5.0pt}
    \resizebox{0.8\linewidth}{!}{
        \begin{tabular}{l|c|c|l}
        \Xhline{1.0pt}
        \multicolumn{1}{c|}{\textbf{\#Query}} &
        \textbf{Instruction} & \textbf{Question} & \multicolumn{1}{c}{\textbf{Avg}} \\
        \Xhline{0.8pt}
        32 + 0 & \cmark & \xmark & 47.8 \\
        32 + 32 & \cmark & \xmark & 50.6 \darkGreen{$\uparrow$2.8} \\
        \rowcolor{gray!20}
        32 + 64 & \cmark & \xmark & \textbf{51.1} \darkGreen{$\uparrow$\textbf{3.3}} \\
        32 + 96 & \cmark & \xmark & 50.7 \darkGreen{$\uparrow$2.9} \\
        \hline
        32 + 64 & \cmark & \cmark & 50.8 \darkGreen{$\uparrow$3.0} \\
        32 + 64 & \xmark & \xmark & 50.5 \darkGreen{$\uparrow$2.7} \\
        \Xhline{1.0pt}
        \end{tabular}
    }
    \vspace{-0.3cm}
    \caption{
    \textbf{QFormer.}
    Introducing more extra queries helps.
    }
    \label{tab:ablation_qformer}
    \vspace{-0.2cm}
\end{table}

%% file: tables/ablation_resolution.tex
\begin{table}[h]
    \centering
    \setlength\tabcolsep{8.0pt}
    \resizebox{0.8\linewidth}{!}{
        \begin{tabular}{l|l|l}
        \Xhline{1.0pt}
        \textbf{Training} &
        \textbf{Testing} & \multicolumn{1}{c}{\textbf{Avg}} \\
        \Xhline{0.8pt}
        \multirow{5}{*}{8$\times$224$\times$224} & 8$\times$224$\times$224 & 50.6 \\
        ~ & 8$\times$\darkGreen{384}$\times$\darkGreen{384} & 49.9 \red{$\downarrow$0.7} \\
        ~ & \cellcolor{gray!20}{\darkGreen{16}$\times$224$\times$224} & \cellcolor{gray!20}{\textbf{51.1} \darkGreen{$\uparrow$\textbf{0.5}}} \\
        ~ & \darkGreen{32}$\times$224$\times$224 & \textbf{51.1} \darkGreen{$\uparrow$\textbf{0.5}} \\
        ~ & \darkGreen{64}$\times$224$\times$224 & 51.0 \darkGreen{$\uparrow$0.4} \\
        \hline
        \darkGreen{16}$\times$224$\times$224 & \darkGreen{16}$\times$224$\times$224 & 51.0 \darkGreen{$\uparrow$0.4} \\
        \Xhline{1.0pt}
        \end{tabular}
    }
    \vspace{-0.3cm}
    \caption{\textbf{Resolution \& Frame.} Large resolution is harmful, while more frames are better for \BenchName.}
    \label{tab:ablation_resolution}
    \vspace{-0.2cm}
\end{table}

%% file: tables/ablation_data2.tex
\begin{table}[h]
    \centering
    \setlength\tabcolsep{4.0pt}
    \resizebox{0.63\linewidth}{!}{
        \begin{tabular}{l|l}
        \Xhline{1.0pt}
        \textbf{Data} & \multicolumn{1}{c}{\textbf{Avg}} \\
        \Xhline{0.8pt}
        \cellcolor{gray!20}{ALL} & \cellcolor{gray!20}{\textbf{51.1}} \\
        ALL $-$ CLEVRER~\cite{clevrer} & 49.3 \red{$\downarrow$1.8} \\
        ALL $-$ SthSthV2~\cite{sth} & 51.0 \red{$\downarrow$0.1} \\
        \Xhline{1.0pt}
        \end{tabular}
    }
    \vspace{-0.3cm}
    \caption{\textbf{Instruction Data.}}
    \label{tab:ablation_data2}
    \vspace{-0.2cm}
\end{table}

%% file: tables/ablation_question_prompt.tex
\begin{table}[h]
    \centering
    \setlength\tabcolsep{4.0pt}
    \resizebox{1.0\linewidth}{!}{
        \begin{tabular}{l|c}
        \Xhline{1.0pt}
        \textbf{Question Prompt} & \textbf{Avg} \\
        \Xhline{0.8pt}
        \rowcolor{gray!20} 
        \textit{Only give the best option.} & \textbf{51.1} \\
        \hline
        \textit{Only give the best option \red{without any explanation.}} & 50.9 \red{$\downarrow$0.2} \\
        \hline
        \textit{\red{Let's think step by step.} Only give the best option.} & 50.5 \red{$\downarrow$0.6} \\
        \Xhline{1.0pt}
        \end{tabular}
    }
    \vspace{-0.3cm}
    \caption{\textbf{Question prompt.} 
    }
    \label{tab:ablation_question_prompt}
    \vspace{-0.2cm}
\end{table}

%% file: tables/nextqa.tex
\begin{table}[tp]
    \centering
    \setlength\tabcolsep{2.0pt}
    \resizebox{1.0\linewidth}{!}{
        \begin{tabular}{l|cccc|cccc}
        \Xhline{1.0pt}
        \multirow{2}{*}{\textbf{Model}} & \multicolumn{4}{c|}{\textbf{Zero-shot}} & \multicolumn{4}{c}{\textbf{In-domain}} \\
        ~ & \textbf{Tem.} & \textbf{Cau.} & \textbf{Des.} & \textbf{Avg} & \textbf{Tem.} & \textbf{Cau.} & \textbf{Des.} & \textbf{Avg}\\
        \Xhline{0.8pt}
        All-in-One~\cite{all_in_one} & - & - & - & - & 48.6 & 48.0 & 63.2 & 50.6 \\
        MIST~\cite{mist} & - & - & - & - & 56.6 & 54.6 & 66.9 & 57.1 \\
        HiTeA~\cite{hitea} & - & - & - & - & 58.3 & 62.4 & 75.6 & 63.1 \\
        InternVideo~\cite{Wang2022InternVideoGV} & 43.4 & 48.0 & 65.1 & 59.1 & 58.3 & 62.4 & 75.6 & 63.1 \\
        \gray{SEVILA~\cite{sevila}} & \gray{61.3} & \gray{61.5} & \gray{75.6} & \gray{63.6} & \gray{69.4} & \gray{74.2} & \gray{81.3}  & \gray{73.8} \\
        \Xhline{0.8pt}
        \textbf{\ModelName} & \textbf{57.4} & \textbf{61.9} & \textbf{69.9} & \textbf{61.7} & 64.7 & 68.7 & 76.1 & 68.6 \\
        \textbf{\ModelName\red{$\dag$}} & - & - & - & - & \textbf{77.0} & \textbf{79.3} & \textbf{79.6} & \textbf{78.6} \\
        \Xhline{1.0pt}
        \end{tabular}
    }
    \vspace{-0.3cm}
    \caption{\textbf{Results on NExT-QA~\cite{nextqa}.}
    ``Tem.'', ``Cau.'' and ``Des.'' stand for ``Temporal'', ``Causal'' and ``Descriptive'' respectively.
    SEVILA~\cite{sevila} is \gray{de-emphasized} since it needs to train an additional localizer.
    For zero-shot results, we remove the NExT-QA in our instruction data.
    ``\red{$\dag$}'' refers to the version with Mistral~\cite{mistral}.
    }
    \label{tab:ft_nextqa}
    \vspace{-0.4cm}
\end{table}

%% file: tables/star_tvqa.tex
\begin{table}[tp]
    \centering
    \setlength\tabcolsep{4.0pt}
    \resizebox{1.0\linewidth}{!}{
        \begin{tabular}{l|ccccc|c}
        \Xhline{1.0pt}
        \multirow{2}{*}{\textbf{Model}} & \multicolumn{5}{c|}{\textbf{STAR}} & \multirow{2}{*}{\textbf{TVQA}} \\
        ~ & \textbf{Int.} & \textbf{Seq.} & \textbf{Pre.}  & \textbf{Fea.} & \textbf{Avg} & ~ \\
        \Xhline{0.8pt}
        FrozenBILM~\cite{frozenbilm} & - & - & - & - & - & 29.7 \\
        InternVideo~\cite{Wang2022InternVideoGV} & 43.8 & 43.2 & 42.3 & 37.4 & 41.6 & 35.9 \\
        \gray{SEVILA~\cite{sevila}} & \gray{48.3} & \gray{45.0} & \gray{44.4}  & \gray{40.8} & \gray{44.6} & \gray{38.2} \\
        \Xhline{0.8pt}
        \textbf{\ModelName} & 58.4 & 60.9 & 55.3 & 53.1 & 59.0 & 40.6 \\
        \textbf{\ModelName\red{$\dag$}} & \textbf{62.4} & \textbf{67.2} & \textbf{57.5} & \textbf{53.9} & \textbf{63.8} & \textbf{46.4} \\
        \Xhline{1.0pt}
        \end{tabular}
    }
    \vspace{-0.3cm}
    \caption{\textbf{Zero-shot results on STAR~\cite{star} and TVQA~\cite{tvqa}.}
    ``Int.'', ``Seq.'', ``Pre.'' and ``Fea.'' stand for ``Interaction'', ``Sequence'', ``Prediction'' and ``Feasibility'' respectively.
    SEVILA~\cite{sevila} is \gray{de-emphasized} since it needs to train an additional localizer.
    For TVQA, we do not input subtitles.
    ``\red{$\dag$}'' refers to the version with Mistral~\cite{mistral}.
    }
    \label{tab:zs_star_tvqa}
    \vspace{-0.4cm}
\end{table}

%% file: tables/egoschema.tex
\begin{table}[tp]
    \centering
    \setlength\tabcolsep{11.0pt}
    \resizebox{1.0\linewidth}{!}{
        \begin{tabular}{l|c|cc}
        \Xhline{1.0pt}
        \multirow{2}{*}{\textbf{Model}} & \multirow{2}{*}{\textbf{Frame}} & \multicolumn{2}{c}{\textbf{EgoSchema}} \\
        ~ & ~ & \textbf{Subset} & \textbf{Fullset} \\
        \Xhline{0.8pt}
        FrozenBILM~\cite{frozenbilm} & 90 & - & 26.9 \\
        VIOLET~\cite{fu2021violet} & 5 & - & 19.9 \\
        mPLUG-Owl~\cite{mplug-owl} & 5 & - & 31.1 \\
        InternVideo~\cite{Wang2022InternVideoGV} & 90 & - & 32.1 \\
        \Xhline{0.8pt}
        \textbf{\ModelName\red{$\dag$}} & 16 & \textbf{63.6} & \textbf{54.4} \\
        \Xhline{1.0pt}
        \end{tabular}
    }
    \vspace{-0.3cm}
    \caption{\textbf{Zero-shot results on EgoSchema~\cite{egoschema}.}
    ``\red{$\dag$}'' refers to the version with Mistral~\cite{mistral}.
    }
    \label{tab:zs_egoschema}
    \vspace{-0.4cm}
\end{table}

%% file: tables/intentqa.tex
\begin{table}[tp]
    \centering
    \setlength\tabcolsep{2.5pt}
    \resizebox{1.0\linewidth}{!}{
        \begin{tabular}{l|cc|cc|cc|cc}
        \Xhline{1.0pt}
        \multirow{2}{*}{\textbf{Model}} & \multicolumn{2}{c|}{\textbf{CW}} & \multicolumn{2}{c|}{\textbf{CH}} & \multicolumn{2}{c|}{\textbf{TP\&TN}} & \multicolumn{2}{c}{\textbf{Total}} \\
        ~ & \textbf{V} & \textbf{T} & \textbf{V} & \textbf{T} & \textbf{V} & \textbf{T} & \textbf{V} & \textbf{T} \\
        \Xhline{0.8pt}
        HQGA~\cite{hqga} & 45.9 & 48.2 & 57.8 & 54.3 & 44.8 & 41.7 & 47.6 & 47.7 \\
        VGT~\cite{vgt} & 50.5 & 51.4 & 56.0 & 56.0 & 48.3 & 47.6 & 50.8 & 51.3 \\
        IntentQA~\cite{intentqa} & - & 58.4 & - & 65.5 & - & 50.5 & - & 57.6 \\
        \rowcolor{gray!20}
        Human & - & 77.8 & - & 80.2 & - & 79.1 & - & 78.5 \\
        \Xhline{0.8pt}
        \textbf{\ModelName\red{$\dag$}} & \textbf{82.5} & \textbf{82.6} & \textbf{86.5} & \textbf{86.9} & \textbf{72.2} & \textbf{77.0} & \textbf{81.9} & \textbf{83.4} \\
        \Xhline{1.0pt}
        \end{tabular}
    }
    \vspace{-0.3cm}
    \caption{\textbf{Results on IntentQA~\cite{intentqa}.}
    ``CW'', ``CH'', ``TP'' and ``TH'' stand for ``Causal Why'', ``Causal How'', ``Temporal Previous'' and ``Temporal Next'' respectively.
    ``V'' and ``T'' stand for ``Validation'' and ``Testing'' split respectively.
    ``\red{$\dag$}'' refers to the version with Mistral~\cite{mistral}.
    }
    \label{tab:ft_intentqa}
    \vspace{-0.4cm}
\end{table}

%% file: tables/bench_details.tex
\begin{table*}[tp]
    \centering
    \setlength\tabcolsep{2pt}
    \resizebox{1.0\textwidth}{!}{
        \begin{tabular}{l|l|l|l|l}
        \Xhline{1.0pt}
        \textbf{Task} & \textbf{Source} & \textbf{Domain} & \textbf{Data Filtration} & \textbf{QA Generation} \\
        \Xhline{1.0pt}
        \textbf{Action Sequence} & STAR~\cite{star} 
        & \makecell[l]{$\cdot$ Real-world\\$\cdot$ Indoor\\$\cdot$ Third-person} 
        & \makecell[l]{
            \cmark\  $\mathrm{Duration} \in(5, 22)$\\
            \cmark\  $\mathrm{Data} \in \mathbf{Prediction}$\\
            \xmark\  $len(\mathrm{A})=1\ \lor \ \mathrm{A}.split(``\  ")=``the"$\\
        } 
        & \makecell[l]{QA: Directly adopt} \\
        \hline
        \textbf{Action Antonym} & PAXION~\cite{paxion}
        & \makecell[l]{$\cdot$ Real-world\&Simulated\\$\cdot$ Indoor\&Outdoor\\$\cdot$ Third-person} 
        & \makecell[l]{N/A} 
        & \makecell[l]{Q: ChatGPT generates\\A: GT$+$Antonym$+$``not sure"} \\
        \hline
        \textbf{Fine-grained Action} & MiT V1~\cite{mit}
        & \makecell[l]{$\cdot$ Real-world\&Simulated\\$\cdot$ Indoor\&Outdoor\\$\cdot$ Third-person} 
        & \makecell[l]{N/A} 
        & \makecell[l]{Q: ChatGPT generates\\A: Randomly sample 4 actions from\\\ \ \ \ \ top-6 predictions of UMT-L/16~\cite{umt}} \\
        \hline
        \textbf{Unexpected Action} & FunQA~\cite{funqa}
        & \makecell[l]{$\cdot$ Real-world\\$\cdot$ Indoor\&Outdoor\\$\cdot$ Third-person} 
        & \makecell[l]{
            \cmark\  $len(\mathrm{QA}\in\mathbf{H2})=34, len(\mathrm{QA}\in\mathbf{H3})=33$\\
            \cmark\  $len(\mathrm{QA}\in\mathbf{C2})=33, len(\mathrm{QA}\in\mathbf{C3})=33$\\
            \cmark\  $len(\mathrm{QA}\in\mathbf{M2})=34, len(\mathrm{QA}\in\mathbf{M3})=33$\\
        } 
        & \makecell[l]{QA: ChatGPT generates from\\\ \ \ \ \ \ \ \ original QA} \\
        \hline
        \textbf{Object Existence} & CLEVRER~\cite{clevrer} 
        & \makecell[l]{$\cdot$ Simulated\\$\cdot$ Indoor} 
        & \makecell[l]{
            \cmark\  $\mathrm{Data} \in \mathbf{desctiptive}\ \land\ \mathrm{Data} \in \mathbf{exist}$\\
            \cmark\  $len(\mathrm{program})<11$\\
        } 
        & \makecell[l]{Q: ChatGPT generates\\A: ``yes''$+$``no''$+$``not sure"} \\
        \hline
        \textbf{Object Interaction} & STAR~\cite{star} 
        & \makecell[l]{$\cdot$ Real-world\\$\cdot$ Indoor\\$\cdot$ Third-person} 
        & \makecell[l]{
            \cmark\  $\mathrm{Duration} \in(7, 20)$\\
            \cmark\  $\mathrm{Data} \in\mathbf{Interation}$\\
            \cmark\  $``object"\ in\ \mathrm{Q}\ \lor\ ``to\ the"\ in\ \mathrm{Q}$\\
        } 
        & \makecell[l]{QA: Directly adopt} \\
        \hline
        \textbf{Object Shuflle} & \makecell[l]{Perception\\Test~\cite{perception_test}}
        & \makecell[l]{$\cdot$ Real-world\\$\cdot$ Indoor\\$\cdot$ First\&Third-person} 
        & \makecell[l]{
            \cmark\  $\mathrm{Data} \in\mathbf{object\ permanence}$\\
            \cmark\  $``Where\ is\ the"\ in\ \mathrm{Q}$\\
        } 
        & \makecell[l]{QA: Directly adopt} \\
        \hline
        \textbf{Moving Direction} & CLEVRER~\cite{clevrer} 
        & \makecell[l]{$\cdot$ Simulated\\$\cdot$ Indoor} 
        & \makecell[l]{
            Select videos where a certain object is either\\stationary or moving in a single direction
        } 
        & \makecell[l]{Q: ChatGPT generates\\A: $\nearrow \searrow \nwarrow \swarrow+$ ``stationary''} \\
        \hline
        \textbf{Action Localization} & \makecell[l]{Charades\\-STA~\cite{charades_sta}} 
        & \makecell[l]{$\cdot$ Real-world\\$\cdot$ Indoor\\$\cdot$ Third-person} 
        & \makecell[l]{
            \cmark\  $\mathrm{Duration}_{entire} > 15$\\
            \cmark\  $\mathrm{Duration}_{start,end,middle} \in (5, 8)$\\
            \xmark\  $``person\ they"\ in\ \mathrm{Q}\ \lor\ ``person\ \ so\ they"\ in\ \mathrm{Q}$\\
        } 
        & \makecell[l]{Q: ChatGPT generates\\A: ``start''$+$``end''$+$``middle''$+$``entire''} \\
        \hline
        \textbf{Scene Transition} & \makecell[l]{MoVQA~\cite{movqa}} 
        & \makecell[l]{$\cdot$ Real-world\\$\cdot$ Indoor\&Outdoor\\$\cdot$ Third-person} 
        & \makecell[l]{Select videos with continuous scene labels} 
        & \makecell[l]{QA: ChatGPT generates from\\\ \ \ \ \ \ \ \ original QA} \\
        \hline
        \textbf{Action Count} & \makecell[l]{Perception\\Test~\cite{perception_test}}
        & \makecell[l]{$\cdot$ Real-world\\$\cdot$ Indoor\\$\cdot$ First\&Third-person} 
        & \makecell[l]{
            \cmark\  $\mathrm{Data} \in\mathbf{action\ counting}$\\
        } 
        & \makecell[l]{QA: Directly adopt} \\
        \hline
        \textbf{Moving Count} & CLEVRER~\cite{clevrer} 
        & \makecell[l]{$\cdot$ Simulated\\$\cdot$ Indoor} 
        & \makecell[l]{
            \cmark\  $\mathrm{Data} \in \mathbf{desctiptive}\ \land\ \mathrm{Data} \in \mathbf{count}$\\
            \cmark\  $len(\mathrm{program})<9$\\
        } 
        & \makecell[l]{Q: ChatGPT generates\\A: Randomly shift original answer} \\
        \hline
        \textbf{Moving Attribute} & CLEVRER~\cite{clevrer} 
        & \makecell[l]{$\cdot$ Simulated\\$\cdot$ Indoor} 
        & \makecell[l]{
            \cmark\  $\mathrm{Data} \in \mathbf{desctiptive}\ \land\ \mathrm{Data} \in \mathbf{query\_color}$\\
            \cmark\  $\mathrm{Data} \in \mathbf{desctiptive}\ \land\ \mathrm{Data} \in \mathbf{query\_shape}$\\
            \cmark\  $\mathrm{Data} \in \mathbf{desctiptive}\ \land\ \mathrm{Data} \in \mathbf{query\_material}$\\
            \cmark\  $len(\mathrm{program})<13$\\
        } 
        & \makecell[l]{Q: ChatGPT generates\\A: Randomly select from candidates} \\
        \hline
        \textbf{State Change} & \makecell[l]{Perception\\Test~\cite{perception_test}}
        & \makecell[l]{$\cdot$ Real-world\\$\cdot$ Indoor\\$\cdot$ First\&Third-person} 
        & \makecell[l]{
            \cmark\  $\mathrm{Data} \in\mathbf{state\ recognition}$\\
            \xmark\  $\mathrm{Q}\  requires\ audio$\\
        } 
        & \makecell[l]{QA: Directly adopt} \\
        \hline
        \textbf{Fine-grained Pose} & \makecell[l]{NTU\\RGB+D~\cite{ntu_rgbd}}
        & \makecell[l]{$\cdot$ Real-world\\$\cdot$ Indoor\\$\cdot$ Third-person} 
        & \makecell[l]{
            Select videos with specific poses
        } 
        & \makecell[l]{Q: ChatGPT generates\\A: Randomly select from similar poses} \\
        \hline
        \textbf{Character Order} & \makecell[l]{Perception\\Test~\cite{perception_test}}
        & \makecell[l]{$\cdot$ Real-world\\$\cdot$ Indoor\\$\cdot$ First\&Third-person} 
        & \makecell[l]{
            \cmark\  $\mathrm{Data} \in\mathbf{letter}$\\
            \cmark\  $``order"\in \mathrm{Q}$\\
        } 
        & \makecell[l]{QA: Directly adopt} \\
        \hline
        \textbf{\makecell[l]{Egocentric\\Navigation}} & \makecell[l]{VLN-CE~\cite{vln_ce}}
        & \makecell[l]{$\cdot$ Simulated\\$\cdot$ Indoor\\$\cdot$ First-person} 
        & \makecell[l]{
            \cmark\  $moving\ forward > 0.75m$\\
            \cmark\  $turning\ left/right \in (60^{\circ}, 120^{\circ})$\\\ \ \ \ $then\ moving\ forward > 0.75m$\\
            \cmark\  $stop$\\
        } 
        & \makecell[l]{Q: ChatGPT generates\\A:``move forward''$+$``stop''\\\ \ \ \ ``turn left and move forward''$+$\\\ \ \ \ ``turn right and move forward''}\\
        \hline
        \textbf{Episodic Reasoning} & \makecell[l]{TVQA~\cite{tvqa}}
        & \makecell[l]{$\cdot$ Real-world\\$\cdot$ Indoor\&Outdoor\\$\cdot$ Third-person} 
        & \makecell[l]{
            \cmark\  $\mathrm{Duration} \in(25, 40)$\\
        } 
        & \makecell[l]{QA: Directly adopt w/o subtitles} \\
        \hline
        \textbf{\makecell[l]{Counterfactual\\Inference}} & CLEVRER~\cite{clevrer} 
        & \makecell[l]{$\cdot$ Simulated\\$\cdot$ Indoor} 
        & \makecell[l]{
            \cmark\  $\mathrm{Data} \in \mathbf{counterfactual}$\\
            \cmark\  $len(\mathrm{program})<8$\\
        } 
        & \makecell[l]{QA: Directly adopt} \\
        \Xhline{1.0pt}
        \end{tabular}
    }
    \vspace{-0.3cm}
    \caption{\textbf{More details about \BenchName\ generation.}}
    \label{tab:ben_details}
    \vspace{-0.3cm}
\end{table*}

%% file: tables/leaderboard.tex
\begin{table*}[tp]
\centering
\begin{minipage}[t]{0.22\textwidth}
    \vspace{0pt}
    \centering
    \setlength\tabcolsep{4.0pt}
    \resizebox{1\linewidth}{!}{
        \begin{tabular}{c|l|c}
        \textbf{Rank} & \multicolumn{1}{c|}{\textbf{Model}} & \textbf{Acc} \\
        \Xhline{1.0pt}
        \rowcolor{blue!20}\textbf{1} & \video \textbf{\red{\ModelName}} & \textbf{66.0}\\
        \rowcolor{blue!12}\textbf{2} & \image \textbf{Otter-I} & \textbf{34.5}\\
        \rowcolor{blue!6}\textbf{3} & \video \textbf{VideoChat} & \textbf{33.5}\\
        4 & \image LLaVA & 28.0 \\
        5 & \video VideoLLaMA & 27.5 \\
        6 & \image mPLUG-Owl-I & 25.0 \\
        7 & \image BLIP2 & 24.5 \\
        8 & \video VideoChatGPT & 23.5 \\
        9 & \image LLaMA-Adapter & 23.0 \\
        10 & \image InstructBLIP & 20.0 \\
        11 & \image MiniGPT-4 & 16.0 \\
        \end{tabular}
    }
    \subcaption{\textit{Action Sequence}}
\end{minipage}
\hfill
\begin{minipage}[t]{0.22\textwidth}
    \vspace{0pt}
    \centering
    \setlength\tabcolsep{4.0pt}
    \resizebox{1\linewidth}{!}{
        \begin{tabular}{c|l|c}
        \textbf{Rank} & \multicolumn{1}{c|}{\textbf{Model}} & \textbf{Acc} \\
        \Xhline{1.0pt}
        \rowcolor{blue!20}\textbf{1} & \video \textbf{\red{\ModelName}} & \textbf{47.5}\\
        \rowcolor{blue!12}\textbf{2} & \image \textbf{LLaVA} & \textbf{39.5}\\
        \rowcolor{blue!6}\textbf{3} & \image \textbf{Otter-I} & \textbf{32.0}\\
        4 & \image BLIP2 & 29.0 \\
        5 & \image LLaMA-Adapter & 28.0 \\
        6 & \video VideoChat & 26.5 \\
        7 & \video VideoChatGPT & 26.0 \\
        8 & \video VideoLLaMA & 25.5 \\
        9 & \image mPLUG-Owl-I & 20.0 \\
        10 & \image MiniGPT-4 & 18.0 \\
        11 & \image InstructBLIP & 16.5 \\
        \end{tabular}
    }
    \subcaption{\textit{Action Prediction}}
\end{minipage}
\hfill
\begin{minipage}[t]{0.22\textwidth}
    \vspace{0pt}
    \centering
    \setlength\tabcolsep{4.0pt}
    \resizebox{1\linewidth}{!}{
        \begin{tabular}{c|l|c}
        \textbf{Rank} & \multicolumn{1}{c|}{\textbf{Model}} & \textbf{Acc} \\
        \Xhline{1.0pt}
        \rowcolor{blue!20}\textbf{1} & \video \textbf{\red{\ModelName}} & \textbf{83.5}\\
        \rowcolor{blue!12}\textbf{2} & \image \textbf{LLaVA} & \textbf{63.0}\\
        \rowcolor{blue!6}\textbf{3} & \video \textbf{VideoChatGPT} & \textbf{62.0}\\
        4 & \video VideoChat & 56.0 \\
        5 & \image LLaMA-Adapter & 51.0 \\
        6 & \video VideoLLaMA & 51.0 \\
        7 & \image InstructBLIP & 46.0 \\
        8 & \image mPLUG-Owl-I & 44.5 \\
        9 & \image Otter-I & 39.5 \\
        10 & \image BLIP2 & 33.5 \\
        11 & \image MiniGPT-4 & 26.0 \\
        \end{tabular}
    }
    \subcaption{\textit{Action Antonym}}
\end{minipage}
\hfill
\begin{minipage}[t]{0.22\textwidth}
    \vspace{0pt}
    \centering
    \setlength\tabcolsep{4.0pt}
    \resizebox{1\linewidth}{!}{
        \begin{tabular}{c|l|c}
        \textbf{Rank} & \multicolumn{1}{c|}{\textbf{Model}} & \textbf{Acc} \\
        \Xhline{1.0pt}
        \rowcolor{blue!20}\textbf{1} & \video \textbf{\red{\ModelName}} & \textbf{49.5}\\
        \rowcolor{blue!12}\textbf{2} & \video \textbf{VideoChat} & \textbf{33.5}\\
        \rowcolor{blue!6}\textbf{3} & \image \textbf{Otter-I} & \textbf{30.5}\\
        4 & \image LLaVA & 30.5 \\
        5 & \image LLaMA-Adapter & 30.0 \\
        6 & \video VideoLLaMA & 29.0 \\
        7 & \image mPLUG-Owl-I & 27.0 \\
        8 & \image InstructBLIP & 24.5 \\
        9 & \video VideoChatGPT & 22.5 \\
        10 & \image MiniGPT-4 & 21.5 \\
        11 & \image BLIP2 & 17.0 \\
        \end{tabular}
    }
    \subcaption{\textit{Fine-grained Action}}
\end{minipage}

\vspace{0.1cm}

\begin{minipage}[t]{0.22\textwidth}
    \vspace{0pt}
    \centering
    \setlength\tabcolsep{4.0pt}
    \resizebox{1\linewidth}{!}{
        \begin{tabular}{c|l|c}
        \textbf{Rank} & \multicolumn{1}{c|}{\textbf{Model}} & \textbf{Acc} \\
        \Xhline{1.0pt}
        \rowcolor{blue!20}\textbf{1} & \video \textbf{\red{\ModelName}} & \textbf{60.0}\\
	\rowcolor{blue!12}\textbf{2} & \image \textbf{InstructBLIP} & \textbf{46.0}\\
	\rowcolor{blue!6}\textbf{3} & \image \textbf{BLIP2} & \textbf{42.0}\\
	4 & \video VideoChat & 40.5 \\
	5 & \image LLaVA & 39.0 \\
	6 & \video VideoLLaMA & 39.0 \\
	7 & \image Otter-I & 38.5 \\
	8 & \image LLaMA-Adapter & 33.0 \\
	9 & \video VideoChatGPT & 26.5 \\
	10 & \image mPLUG-Owl-I & 23.5 \\
	11 & \image MiniGPT-4 & 16.0 \\
        \end{tabular}
    }
    \subcaption{\textit{Unexpected Action}}
\end{minipage}
\hfill
\begin{minipage}[t]{0.22\textwidth}
    \vspace{0pt}
    \centering
    \setlength\tabcolsep{3.pt}
    \resizebox{1\linewidth}{!}{
        \begin{tabular}{c|l|c}
        \textbf{Rank} & \multicolumn{1}{c|}{\textbf{Model}} & \textbf{Acc} \\
        \Xhline{1.0pt}
        \rowcolor{blue!20}\textbf{1} & \video \textbf{\red{\ModelName}} & \textbf{58.0}\\
	\rowcolor{blue!12}\textbf{2} & \video \textbf{VideoChatGPT} & \textbf{54.0}\\
	\rowcolor{blue!6}\textbf{3} & \image \textbf{LLaMA-Adapter} & \textbf{53.5}\\
	4 & \image LLaVA & 53.0 \\
	5 & \video VideoChat & 53.0 \\
	6 & \image BLIP2 & 51.5 \\
	7 & \image InstructBLIP & 51.0 \\
	8 & \image Otter-I & 48.5 \\
	9 & \video VideoLLaMA & 48.0 \\
	10 & \image mPLUG-Owl-I & 36.0 \\
	11 & \image MiniGPT-4 & 29.5 \\
        \end{tabular}
    }
    \subcaption{\textit{Object Existence}}
\end{minipage}
\hfill
\begin{minipage}[t]{0.22\textwidth}
    \vspace{0pt}
    \centering
    \setlength\tabcolsep{4.0pt}
    \resizebox{1\linewidth}{!}{
        \begin{tabular}{c|l|c}
        \textbf{Rank} & \multicolumn{1}{c|}{\textbf{Model}} & \textbf{Acc} \\
        \Xhline{1.0pt}
        \rowcolor{blue!20}\textbf{1} & \video \textbf{\red{\ModelName}} & \textbf{71.5}\\
	\rowcolor{blue!12}\textbf{2} & \image \textbf{Otter-I} & \textbf{44.0}\\
	\rowcolor{blue!6}\textbf{3} & \image \textbf{LLaVA} & \textbf{41.0}\\
	4 & \video VideoLLaMA & 40.5 \\
	5 & \video VideoChat & 40.5 \\
	6 & \image LLaMA-Adapter & 32.5 \\
	7 & \video VideoChatGPT & 28.0 \\
	8 & \image BLIP2 & 26.0 \\
	9 & \image InstructBLIP & 26.0 \\
	10 & \image MiniGPT-4 & 25.5 \\
	11 & \image mPLUG-Owl-I & 24.0 \\
        \end{tabular}
    }
    \subcaption{\textit{Object Interaction}}
\end{minipage}
\hfill
\begin{minipage}[t]{0.22\textwidth}
    \vspace{0pt}
    \centering
    \setlength\tabcolsep{4.0pt}
    \resizebox{1\linewidth}{!}{
        \begin{tabular}{c|l|c}
        \textbf{Rank} & \multicolumn{1}{c|}{\textbf{Model}} & \textbf{Acc} \\
        \Xhline{1.0pt}
        \rowcolor{blue!20}\textbf{1} & \video \textbf{\red{\ModelName}} & \textbf{42.5}\\
	\rowcolor{blue!12}\textbf{2} & \image \textbf{LLaVA} & \textbf{41.5}\\
	\rowcolor{blue!6}\textbf{3} & \video \textbf{VideoChatGPT} & \textbf{40.0}\\
	4 & \video VideoLLaMA & 38.0 \\
	5 & \image InstructBLIP & 37.5 \\
	6 & \image mPLUG-Owl-I & 34.0 \\
	7 & \image LLaMA-Adapter & 33.5 \\
	8 & \image BLIP2 & 31.0 \\
	9 & \video VideoChat & 30.0 \\
	10 & \image Otter-I & 29.5 \\
	11 & \image MiniGPT-4 & 13.0 \\

        \end{tabular}
    }
    \subcaption{\textit{Object Shuffle}}
\end{minipage}

\vspace{0.1cm}

\begin{minipage}[t]{0.22\textwidth}
    \vspace{0pt}
    \centering
    \setlength\tabcolsep{3.0pt}
    \resizebox{1\linewidth}{!}{
        \begin{tabular}{c|l|c}
        \textbf{Rank} & \multicolumn{1}{c|}{\textbf{Model}} & \textbf{Acc} \\
        \Xhline{1.0pt}
        \rowcolor{blue!20}\textbf{1} & \image \textbf{LLaMA-Adapter} & \textbf{25.5}\\
	\rowcolor{blue!12}\textbf{2} & \image \textbf{BLIP2} & \textbf{25.5}\\
	\rowcolor{blue!6}\textbf{3} & \video \textbf{VideoChat} & \textbf{25.5}\\
	4 & \video \red{\ModelName} & 23.0 \\
	5 & \video VideoChatGPT & 23.0 \\
	6 & \image mPLUG-Owl-I & 23.0 \\
	7 & \image LLaVA & 23.0 \\
	8 & \video VideoLLaMA & 22.5 \\
	9 & \image InstructBLIP & 22.0 \\
	10 & \image Otter-I & 19.0 \\
	11 & \image MiniGPT-4 & 11.5 \\
        \end{tabular}
    }
    \subcaption{\textit{Moving Direction}}
\end{minipage}
\hfill
\begin{minipage}[t]{0.22\textwidth}
    \vspace{0pt}
    \centering
    \setlength\tabcolsep{4.0pt}
    \resizebox{1\linewidth}{!}{
        \begin{tabular}{c|l|c}
        \textbf{Rank} & \multicolumn{1}{c|}{\textbf{Model}} & \textbf{Acc} \\
        \Xhline{1.0pt}
        \rowcolor{blue!20}\textbf{1} & \video \textbf{VideoChat} & \textbf{27.0}\\
	\rowcolor{blue!12}\textbf{2} & \image \textbf{BLIP2} & \textbf{26.0}\\
	\rowcolor{blue!6}\textbf{3} & \image \textbf{Otter-I} & \textbf{25.5}\\
	4 & \image mPLUG-Owl-I & 24.0 \\
	5 & \video \red{\ModelName} & 23.0 \\
	6 & \image InstructBLIP & 23.0 \\
	7 & \video VideoLLaMA & 22.5 \\
	8 & \image LLaMA-Adapter & 21.5 \\
	9 & \image LLaVA & 20.5 \\
	10 & \video VideoChatGPT & 20.0 \\
	11 & \image MiniGPT-4 & 12.0 \\
        \end{tabular}
    }
    \subcaption{\textit{Action Localization}}
\end{minipage}
\hfill
\begin{minipage}[t]{0.22\textwidth}
    \vspace{0pt}
    \centering
    \setlength\tabcolsep{4.0pt}
    \resizebox{1\linewidth}{!}{
        \begin{tabular}{c|l|c}
        \textbf{Rank} & \multicolumn{1}{c|}{\textbf{Model}} & \textbf{Acc} \\
        \Xhline{1.0pt}
        \rowcolor{blue!20}\textbf{1} & \video \textbf{\red{\ModelName}} & \textbf{88.5}\\
	\rowcolor{blue!12}\textbf{2} & \image \textbf{Otter-I} & \textbf{55.0}\\
	\rowcolor{blue!6}\textbf{3} & \video \textbf{VideoChat} & \textbf{48.5}\\
	4 & \image InstructBLIP & 46.5 \\
	5 & \image LLaVA & 45.0 \\
	6 & \video VideoLLaMA & 43.0 \\
	7 & \image mPLUG-Owl-I & 34.5 \\
	8 & \image BLIP2 & 32.5 \\
	9 & \video VideoChatGPT & 31.0 \\
	10 & \image LLaMA-Adapter & 30.5 \\
	11 & \image MiniGPT-4 & 9.5 \\
        \end{tabular}
    }
    \subcaption{\textit{Scene transition}}
\end{minipage}
\hfill
\begin{minipage}[t]{0.22\textwidth}
    \vspace{0pt}
    \centering
    \setlength\tabcolsep{4.0pt}
    \resizebox{1\linewidth}{!}{
        \begin{tabular}{c|l|c}
        \textbf{Rank} & \multicolumn{1}{c|}{\textbf{Model}} & \textbf{Acc} \\
        \Xhline{1.0pt}
        \rowcolor{blue!20}\textbf{1} & \image \textbf{InstructBLIP} & \textbf{42.5}\\
	\rowcolor{blue!12}\textbf{2} & \video \textbf{\red{\ModelName}} & \textbf{39.0}\\
	\rowcolor{blue!6}\textbf{3} & \video \textbf{VideoChat} & \textbf{35.0}\\
	4 & \image mPLUG-Owl-I & 34.5 \\
	5 & \image LLaVA & 34.0 \\
	6 & \video VideoLLaMA & 34.0 \\
	7 & \image MiniGPT-4 & 32.5 \\
	8 & \video VideoChatGPT & 30.5 \\
	9 & \image LLaMA-Adapter & 29.0 \\
	10 & \image BLIP2 & 25.5 \\
	11 & \image Otter-I & 20.0 \\
        \end{tabular}
    }
    \subcaption{\textit{Action Count}}
\end{minipage}

\vspace{0.1cm}

\begin{minipage}[t]{0.22\textwidth}
    \vspace{0pt}
    \centering
    \setlength\tabcolsep{4.0pt}
    \resizebox{1\linewidth}{!}{
        \begin{tabular}{c|l|c}
        \textbf{Rank} & \multicolumn{1}{c|}{\textbf{Model}} & \textbf{Acc} \\
        \Xhline{1.0pt}
        \rowcolor{blue!20}\textbf{1} & \video \textbf{\red{\ModelName}} & \textbf{42.0}\\
	\rowcolor{blue!12}\textbf{2} & \image \textbf{Otter-I} & \textbf{32.5}\\
	\rowcolor{blue!6}\textbf{3} & \image \textbf{BLIP2} & \textbf{30.0}\\
	4 & \image InstructBLIP & 26.5 \\
	5 & \video VideoChatGPT & 25.5 \\
	6 & \video VideoLLaMA & 22.5 \\
	7 & \image LLaMA-Adapter & 22.5 \\
	8 & \image mPLUG-Owl-I & 22.0 \\
	9 & \image LLaVA & 20.5 \\
	10 & \video VideoChat & 20.5 \\
	11 & \image MiniGPT-4 & 15.5 \\
        \end{tabular}
    }
    \subcaption{\textit{Moving Count}}
\end{minipage}
\hfill
\begin{minipage}[t]{0.22\textwidth}
    \vspace{0pt}
    \centering
    \setlength\tabcolsep{4.0pt}
    \resizebox{1\linewidth}{!}{
        \begin{tabular}{c|l|c}
        \textbf{Rank} & \multicolumn{1}{c|}{\textbf{Model}} & \textbf{Acc} \\
        \Xhline{1.0pt}
        \rowcolor{blue!20}\textbf{1} & \video \textbf{VideoChatGPT} & \textbf{48.5}\\
	\rowcolor{blue!12}\textbf{2} & \image \textbf{LLaVA} & \textbf{47.0}\\
	\rowcolor{blue!6}\textbf{3} & \video \textbf{VideoChat} & \textbf{46.0}\\
	4 & \video VideoLLaMA & 45.5 \\
	5 & \video \red{\ModelName} & 44.0 \\
	6 & \image BLIP2 & 42.0 \\
	7 & \image mPLUG-Owl-I & 40.0 \\
	8 & \image LLaMA-Adapter & 39.5 \\
	9 & \image Otter-I & 39.0 \\
	10 & \image MiniGPT-4 & 34.0 \\
	11 & \image InstructBLIP & 32.0 \\
        \end{tabular}
    }
    \subcaption{\textit{Moving Attribute}}
\end{minipage}
\hfill
\begin{minipage}[t]{0.22\textwidth}
    \vspace{0pt}
    \centering
    \setlength\tabcolsep{4.0pt}
    \resizebox{1\linewidth}{!}{
        \begin{tabular}{c|l|c}
        \textbf{Rank} & \multicolumn{1}{c|}{\textbf{Model}} & \textbf{Acc} \\
        \Xhline{1.0pt}
        \rowcolor{blue!20}\textbf{1} & \video \textbf{\red{\ModelName}} & \textbf{49.0}\\
	\rowcolor{blue!12}\textbf{2} & \video \textbf{VideoLLaMA} & \textbf{32.5}\\
	\rowcolor{blue!6}\textbf{3} & \video \textbf{VideoChatGPT} & \textbf{29.0}\\
	4 & \image Otter-I & 28.0 \\
	5 & \image BLIP2 & 27.0 \\
	6 & \video VideoChat & 26.5 \\
	7 & \image MiniGPT-4 & 26.0 \\
	8 & \image InstructBLIP & 25.5 \\
	9 & \image LLaMA-Adapter & 25.0 \\
	10 & \image LLaVA & 25.0 \\
	11 & \image mPLUG-Owl-I & 24.0 \\
        \end{tabular}
    }
    \subcaption{\textit{State Change}}
\end{minipage}
\hfill
\begin{minipage}[t]{0.22\textwidth}
    \vspace{0pt}
    \centering
    \setlength\tabcolsep{3.0pt}
    \resizebox{1\linewidth}{!}{
        \begin{tabular}{c|l|c}
        \textbf{Rank} & \multicolumn{1}{c|}{\textbf{Model}} & \textbf{Acc} \\
        \Xhline{1.0pt}
        \rowcolor{blue!20}\textbf{1} & \video \textbf{\red{\ModelName}} & \textbf{58.5}\\
	\rowcolor{blue!12}\textbf{2} & \video \textbf{VideoChat} & \textbf{42.5}\\
	\rowcolor{blue!6}\textbf{3} & \image \textbf{LLaMA-Adapter} & \textbf{41.5}\\
	4 & \image InstructBLIP & 40.5 \\
	5 & \image BLIP2 & 40.0 \\
	6 & \video VideoChatGPT & 39.5 \\
	7 & \image LLaVA & 38.5 \\
	8 & \video VideoLLaMA & 32.5 \\
	9 & \image mPLUG-Owl-I & 31.5 \\
	10 & \image Otter-I & 28.5 \\
	11 & \image MiniGPT-4 & 8.0 \\
        \end{tabular}
    }
    \subcaption{\textit{Fine-grained Pose}}
\end{minipage}

\vspace{0.1cm}

\begin{minipage}[t]{0.22\textwidth}
    \vspace{0pt}
    \centering
    \setlength\tabcolsep{4.0pt}
    \resizebox{1\linewidth}{!}{
        \begin{tabular}{c|l|c}
        \textbf{Rank} & \multicolumn{1}{c|}{\textbf{Model}} & \textbf{Acc} \\
        \Xhline{1.0pt}
        \rowcolor{blue!20}\textbf{1} & \video \textbf{VideoChat} & \textbf{41.0}\\
	\rowcolor{blue!12}\textbf{2} & \video \textbf{VideoLLaMA} & \textbf{40.0}\\
	\rowcolor{blue!6}\textbf{3} & \image \textbf{mPLUG-Owl-I} & \textbf{37.0}\\
	4 & \video \red{\ModelName} & 36.5 \\
	5 & \image LLaVA & 36.0 \\
	6 & \video VideoChatGPT & 33.0 \\
	7 & \image LLaMA-Adapter & 31.5 \\
	8 & \image BLIP2 & 30.0 \\
	9 & \image InstructBLIP & 30.0 \\
	10 & \image MiniGPT-4 & 29.5 \\
	11 & \image Otter-I & 27.0 \\
        \end{tabular}
    }
    \subcaption{\textit{Character Order}}
\end{minipage}
\hfill
\begin{minipage}[t]{0.22\textwidth}
    \vspace{0pt}
    \centering
    \setlength\tabcolsep{4.0pt}
    \resizebox{1\linewidth}{!}{
        \begin{tabular}{c|l|c}
        \textbf{Rank} & \multicolumn{1}{c|}{\textbf{Model}} & \textbf{Acc} \\
        \Xhline{1.0pt}
        \rowcolor{blue!20}\textbf{1} & \video \textbf{\red{\ModelName}} & \textbf{35.0}\\
	\rowcolor{blue!12}\textbf{2} & \image \textbf{Otter-I} & \textbf{32.0}\\
	\rowcolor{blue!6}\textbf{3} & \video \textbf{VideoLLaMA} & \textbf{30.0}\\
	4 & \video VideoChatGPT & 29.5 \\
	5 & \image LLaVA & 27.0 \\
	6 & \image BLIP2 & 26.0 \\
	7 & \image mPLUG-Owl-I & 25.5 \\
	8 & \image InstructBLIP & 25.5 \\
	9 & \video VideoChat & 23.5 \\
	10 & \image LLaMA-Adapter & 22.5 \\
	11 & \image MiniGPT-4 & 19.0 \\
        \end{tabular}
    }
    \subcaption{\textit{Egocentric Navigation}}
\end{minipage}
\hfill
\begin{minipage}[t]{0.22\textwidth}
    \vspace{0pt}
    \centering
    \setlength\tabcolsep{4.0pt}
    \resizebox{1\linewidth}{!}{
        \begin{tabular}{c|l|c}
        \textbf{Rank} & \multicolumn{1}{c|}{\textbf{Model}} & \textbf{Acc} \\
        \Xhline{1.0pt}
        \rowcolor{blue!20}\textbf{1} & \video \textbf{\red{\ModelName}} & \textbf{40.5}\\
	\rowcolor{blue!12}\textbf{2} & \image \textbf{BLIP2} & \textbf{37.0}\\
	\rowcolor{blue!6}\textbf{3} & \image \textbf{InstructBLIP} & \textbf{30.5}\\
	4 & \image Otter-I & 29.0 \\
	5 & \image LLaMA-Adapter & 28.0 \\
	6 & \image LLaVA & 26.5 \\
	7 & \video VideoChatGPT & 26.0 \\
	8 & \video VideoChat & 23.5 \\
	9 & \image mPLUG-Owl-I & 21.0 \\
	10 & \video VideoLLaMA & 21.0 \\
	11 & \image MiniGPT-4 & 9.9 \\
        \end{tabular}
    }
    \subcaption{\textit{Episodic Reasoning}}
\end{minipage}
\hfill
\begin{minipage}[t]{0.22\textwidth}
    \vspace{0pt}
    \centering
    \setlength\tabcolsep{4.0pt}
    \resizebox{1\linewidth}{!}{
        \begin{tabular}{c|l|c}
        \textbf{Rank} & \multicolumn{1}{c|}{\textbf{Model}} & \textbf{Acc} \\
        \Xhline{1.0pt}
        \rowcolor{blue!20}\textbf{1} & \video \textbf{\red{\ModelName}} & \textbf{65.5}\\
	\rowcolor{blue!12}\textbf{2} & \image \textbf{LLaVA} & \textbf{42.0}\\
	\rowcolor{blue!6}\textbf{3} & \image \textbf{InstructBLIP} & \textbf{38.0}\\
	4 & \image mPLUG-Owl-I & 37.0 \\
	5 & \video VideoLLaMA & 37.0 \\
	6 & \image Otter-I & 36.5 \\
	7 & \video VideoChat & 36.0 \\
	8 & \video VideoChatGPT & 35.5 \\
	9 & \image LLaMA-Adapter & 32.0 \\
	10 & \image BLIP2 & 31.0 \\
	11 & \image MiniGPT-4 & 3.0 \\
        \end{tabular}
    }
    \subcaption{\textit{Counterfactual Inference}}
\end{minipage}
\caption{\textbf{Leaderboards of different tasks in \BenchName\ (until \red{2023/11/28}).} Our \ModelName\  secures the top ranking on \textbf{15} tasks. Full results on MVBench can be found at \url{https://huggingface.co/spaces/OpenGVLab/MVBench_Leaderboard}.}
\vspace{-0.3cm}
\label{tab:leaderboard}
\end{table*}